\documentclass{article}

\usepackage{arxiv}

\usepackage[utf8]{inputenc} 
\usepackage[T1]{fontenc}    
\usepackage{hyperref}       
\usepackage{url}            
\usepackage{booktabs}      
\usepackage{amsfonts}       
\usepackage{nicefrac}       
\usepackage{microtype}      
\usepackage{lipsum}
\usepackage{graphicx}
\usepackage[ruled,vlined,linesnumbered]{algorithm2e}

\usepackage{amsmath,amssymb} 
\usepackage{color}
\usepackage{eso-pic}
\usepackage{biblatex}
\usepackage[table]{xcolor}  
\usepackage{booktabs}      
\usepackage{multirow}      
\usepackage{chngcntr}
\usepackage{cleveref}      

\counterwithout{table}{section}
\counterwithout{table}{subsection}
\counterwithout{table}{subsubsection}

\graphicspath{ {./images/} }

\title{VeloEdit: Training-Free Consistent and Continuous Instruction-Based Image Editing via Velocity Field Decomposition}

\author{
 Zongqing Li\\
  Xiamen University,\\ 
  Truesight\\
   \And
 Zhihui Liu \\
  Truesight\\
  \And
 Yujie Xie \\
  Truesight\\
  \And
 Shansiyuan Wu \\
  Truesight\\
  \And
 Hongshen Lv \\
  Xiamen University\\
  \And
 Songzhi Su \\
  Xiamen University\\
}

\begin{document}
\maketitle

\begin{figure}[h]
    \centering 
    \includegraphics[width=0.85\textwidth]{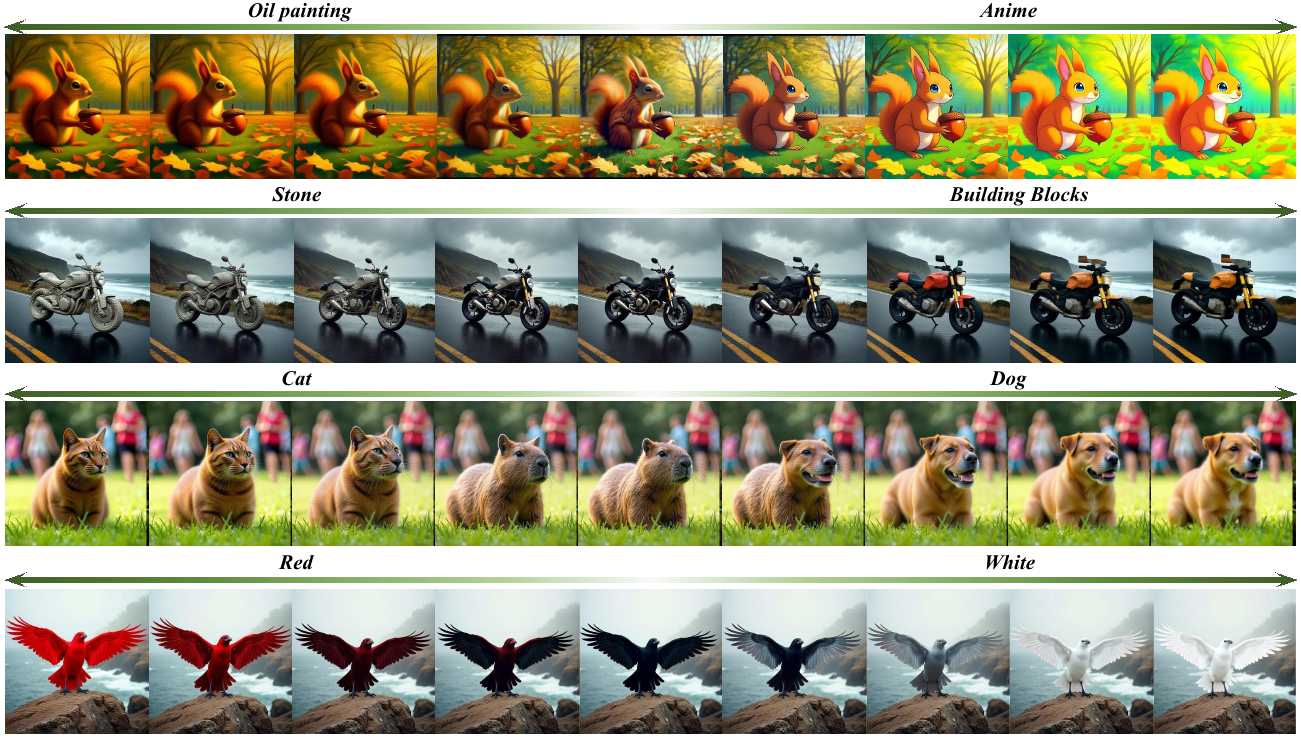}
    \caption{VeloEdit constructs continuous editing trajectories for instruction-based image editing models. Our method empowers these models to achieve continuous and consistent control over edit effects without additional training.} 
    \label{fig:f_title} 
\end{figure}

\begin{abstract}

\label{abs}
  Instruction-based image editing aims to modify source content according to textual instructions. However, existing methods built upon flow matching often struggle to maintain consistency in non-edited regions due to denoising-induced reconstruction errors that cause drift in preserved content. Moreover, they typically lack fine-grained control over edit strength. To address these limitations, we propose VeloEdit, a training-free method that enables highly consistent and continuously controllable editing. VeloEdit dynamically identifies editing regions by quantifying the discrepancy between the velocity fields responsible for preserving source content and those driving the desired edits. Based on this partition, we enforce consistency in preservation regions by substituting the editing velocity with the source-restoring velocity, while enabling continuous modulation of edit intensity in target regions via velocity interpolation. Unlike prior works that rely on complex attention manipulation or auxiliary trainable modules, VeloEdit operates directly on the velocity fields. Extensive experiments on Flux.1 Kontext and Qwen-Image-Edit demonstrate that VeloEdit improves visual consistency and editing continuity with negligible additional computational cost. Code is available at \url{https://github.com/xmulzq/VeloEdit}.
\end{abstract}

\section{Introduction}
\label{sec:intro}

In recent years, diffusion and flow matching models~\cite{ddim,ddpm,ncsn,scoreSDE,Flow,RectifiedFlow} have achieved rapid advancements in generative tasks, showing remarkable progress across diverse domains, including image synthesis~\cite{ddim,scoreSDE,SD}, video generation~\cite{Opensora,wan}, 3D generation~\cite{Dreamfusion,Dreammesh}, and audio synthesis~\cite{audioDiff,Diffsound}. The emergence of large-scale text-to-image models~\cite{SD,SD3,Glide,Imagen,Dalle} has further enhanced the capability to comprehend user intent and improve output controllability, catalyzing the development of instruction-based image editing methods. These methods enable precise editing solely through textual instructions, allowing users to generate high-quality results with a minimal learning curve. However, methods relying exclusively on textual instructions often struggle to preserve consistency in non-edited regions and fail to achieve continuous editing effects. Consequently, the generated outcomes are confined to a limited subset of the model's latent capabilities, restricting their practical application. For instance, given an image of a woman with long hair and the instruction ``dye her hair red'', existing models typically yield outputs with a fixed color intensity, often accompanied by unintended alterations to facial features or background drift.

To further advance the capabilities of large-scale editing models, several methods incorporate source feature maps and editing masks to enhance editing consistency~\cite{kvEdit,ProEdit,follow,SpotEdit}. Furthermore, other studies introduce trainable neural networks~\cite{ConceptSliders,Alchemist,KontinuousKontext} to parameterize editing intensity as a controllable slider. However, these methods typically necessitate extracting feature maps from the source image to derive editing masks, manipulating internal attention computations, or relying on additional training data and computational resources.

\begin{figure}[tbp] 
    \centering 
    \includegraphics[width=0.9\textwidth]{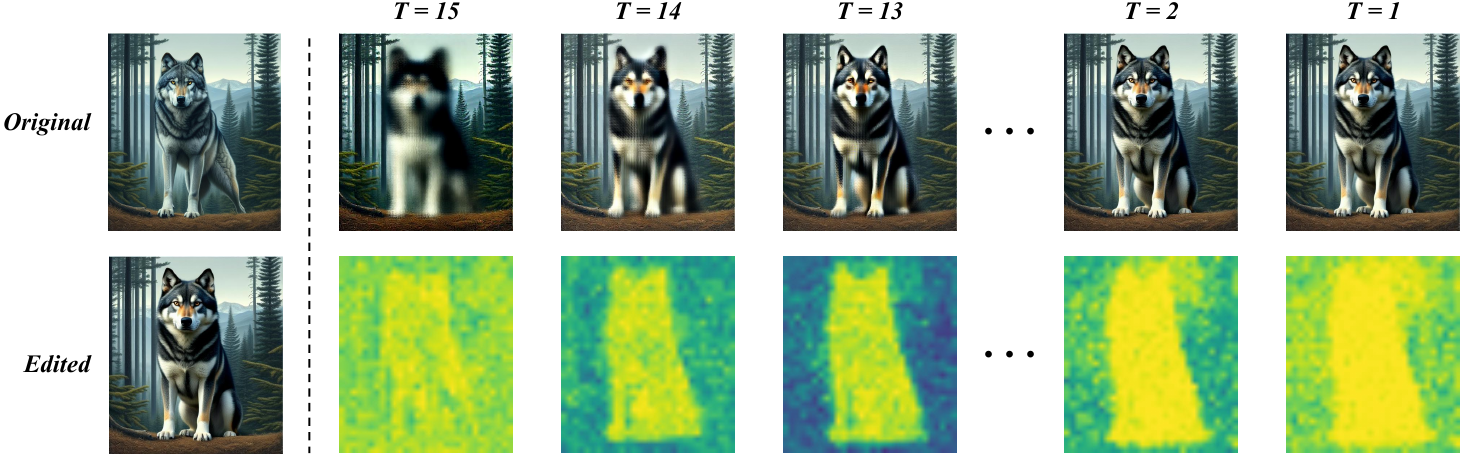}
    \caption{Masks derived via the velocity field. These masks separate preservation regions from editing regions, serving as the foundation for enhancing consistency and enabling precise control over editing intensity.} 
    \label{fig:f_1} 
\end{figure}

This raises a fundamental question regarding large-scale editing models: what are the essential factors that hinder their ability to achieve consistent and continuous editing? Is it strictly necessary to manipulate internal attention mechanisms or introduce auxiliary training modules to unlock these capabilities? We argue that the answer is negative, and large-scale editing models inherently possess these capabilities. However, they are constrained by information loss and feature entanglement within the latent space, alongside the absence of variable-intensity instruction encoding. To explore how to fully unleash this potential, we first visualize the decoded representations of the predicted clean data at each timestep. As illustrated in \cref{fig:f_1}, we observe that as early as the first step, the model has already localized the editing target, while the structure of non-edited regions is substantially established. Subsequent steps primarily focus on restoring non-edited content and refining high-frequency editing details. Furthermore, we experiment with substituting the editing velocity of the first $N$ timesteps with the preservation velocity. We find that intervening in merely the initial one or two timesteps is sufficient to completely suppress the editing effect (see \cref{fig:f_2}). This observation corroborates our hypothesis: the core editing transformation is predominantly governed by the trajectory's initial phase.

\begin{figure}[tbp]
    \centering 
    \includegraphics[width=0.9\textwidth]{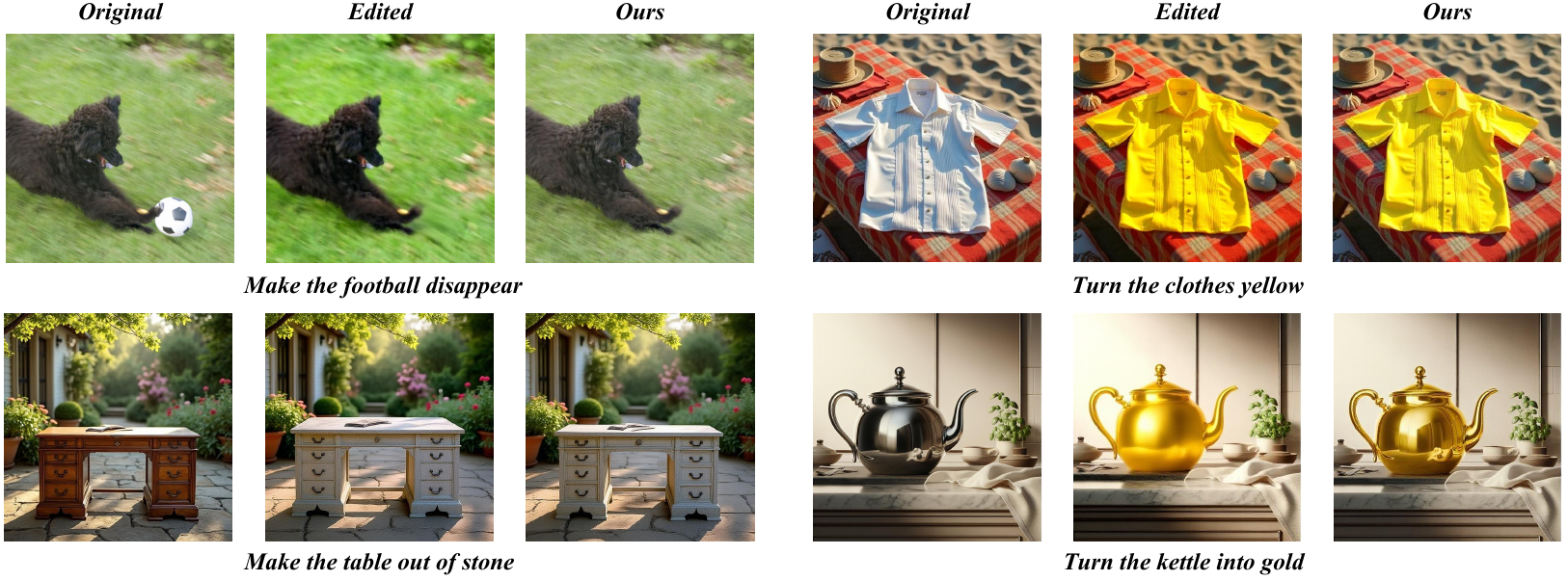}
    \caption{Editing results of the high similarity velocity replacement strategy. By substituting predicted velocities in high similarity regions($S_t > \tau$) with the preservation velocity, VeloEdit effectively maintains the structural integrity of non-edited regions.} 
    \label{fig:f_3} 
\end{figure}

Driven by these findings, we propose VeloEdit, a training-free method that significantly improves the consistency of large-scale editing models and facilitates continuous editing control. Recognizing that global velocity intervention completely nullifies editing effects, we investigate the potential of spatially selective intervention. We first quantify the alignment between the preservation and editing velocity fields. Regions exhibiting high similarity (exceeding a threshold $\tau$) are identified as preservation zones, where we override the editing velocity with the source preservation velocity. This strategy enforces background consistency without compromising the target edit (see \cref{fig:f_3}). Conversely, we posit that regions falling below $\tau$ represent the active editing field driving the semantic transformation. For these regions, we enable continuous editing by modulating the velocity through continuous interpolation and extrapolation between the editing velocity and the preservation velocity. As demonstrated in \cref{fig:exp_img_all}, VeloEdit successfully generates smooth, continuous editing trajectories. In summary, our contributions are outlined as follows:

\begin{itemize} 
    \item We reveal that the editing outcome is predominantly governed by velocity fields in the initial timesteps, whereas subsequent steps focus on preserving non-edited content and refining high-frequency details. 
    \item We propose VeloEdit, a training-free method operating on velocity fields to enhance consistency and enable continuous editing. By modulating velocity, our method ensures visual coherence while achieving continuous, smooth, and fine-grained control over editing intensity.
    \item We validate VeloEdit on large-scale models, including Flux.1 Kontext and Qwen-Image-Edit. Extensive experiments demonstrate superior performance in maintaining visual consistency and editing continuity, confirming the efficacy of VeloEdit. 
\end{itemize}

\section{Related Work}
\label{sec:RelatedWork}
\subsection{Instruction-Based Image Editing}

\begin{figure}[tbp]
    \centering 
    \includegraphics[width=0.9\textwidth]{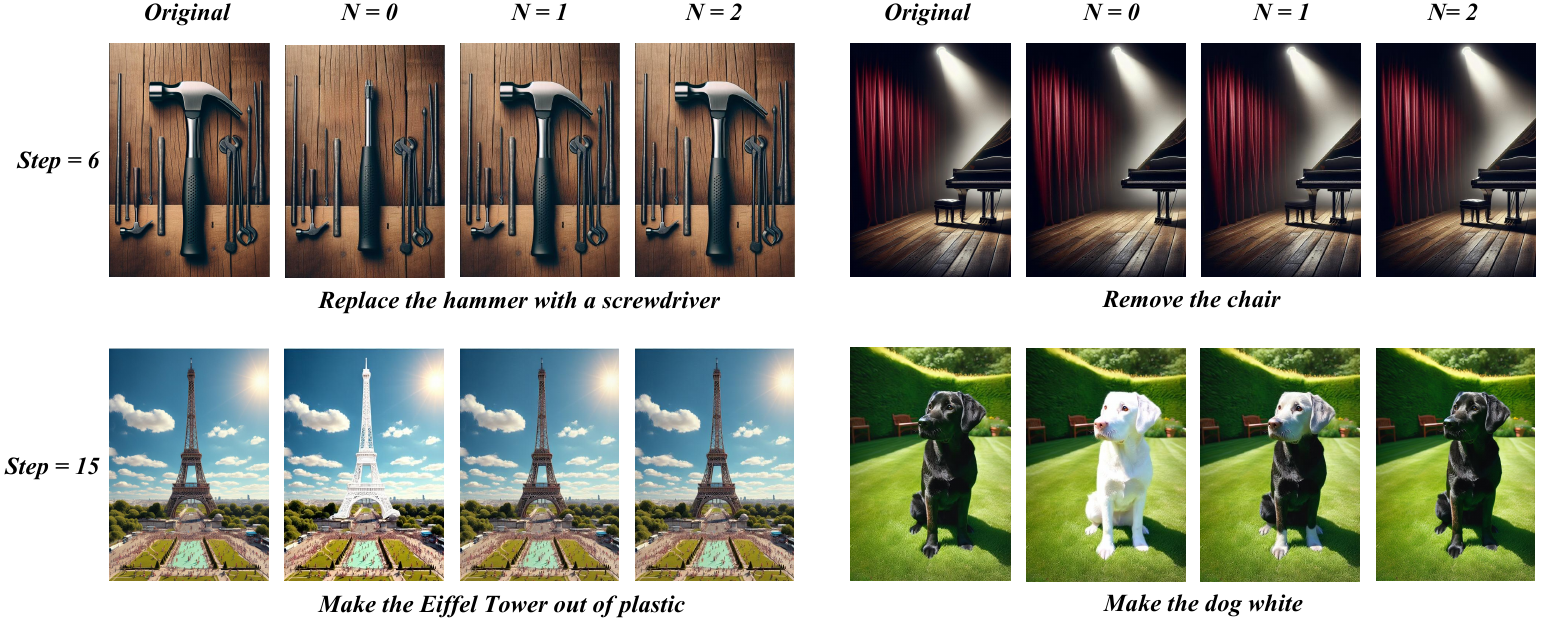}
    \caption{Impact of early velocity replacement. Intervening in the initial one or two timesteps completely suppresses the editing effect.} 
    \label{fig:f_2} 
\end{figure}

Text-to-image (T2I) synthesis models~\cite{SD,SD3,Imagen,Glide,Dalle} have witnessed transformative progress, catalyzing a wide spectrum of sophisticated image editing applications~\cite{ptp,Step1x,FluxKontext,Qwen-Image-Layered,QwenImage,controlNet,Instructpix2pix,IpAdapter}. A pioneering work, Prompt-to-Prompt~\cite{ptp}, facilitates editing by repurposing T2I models through the manipulation of internal cross-attention layers and the injection of source-domain attention maps. This research trajectory has inspired numerous follow-up studies that achieve controllable editing via inversion and attention modulation~\cite{inversion,DiffEdit}. Recent endeavors have further sought to bypass the computationally expensive inversion process by identifying efficient transport paths between source and target distributions~\cite{Flowedit,dvrf}. Additionally, InstructPix2Pix~\cite{Instructpix2pix} leverages Large Language Models (LLMs) and Prompt-to-Prompt for automated data generation, constructing and curating a large-scale triplet dataset consisting of source images, instructions, and edited targets to train a dedicated instruction-based editing model. Following the InstructPix2Pix paradigm, several extensions have introduced auxiliary feature channels to integrate guidance signals~\cite{IpAdapter,controlNet,Step1x}, thereby substantially bolstering the fidelity and controllability of synthesized results.

Recently, the emergence of large-scale image editing models, such as Flux.1 Kontext~\cite{FluxKontext} and Qwen-Image-Edit~\cite{QwenImage}, has markedly expanded the capabilities of instruction-based editing models~\cite{SeedEdit3.0,Qwen-Image-Layered}. By supporting precise editing across diverse tasks solely through textual instructions, these models enable users to generate high-quality results with minimal expertise. However, exclusive reliance on textual guidance poses inherent challenges in maintaining visual consistency and achieving continuous editing effects. This limitation confines editing outcomes to a narrow subset of the model's latent generative capabilities, thereby severely impeding the application potential and creative versatility of large-scale editing models.

\subsection{Consistent Editing}

To improve editing fidelity, prevailing methods leverage inversion techniques to extract intermediate features, specifically Key-Value pairs, from the source image. These features are then injected into the corresponding timesteps of the denoising process, explicitly propagating structural and semantic layout from the source image to the edited output~\cite{kvEdit,ProEdit,follow,SpotEdit}. Alternatively, other methods introduce mask-based mechanisms to disentangle editable regions from the background. By integrating feature injection with masking, these strategies effectively shield the preservation regions from unintended modifications during the denoising process~\cite{SpotEdit,DiffEdit,Magicbrush,Ominicontrol2,RegionE}

In contrast, our method bypasses the cumbersome inversion and KV feature injection procedures, and directly intervenes in the velocity field. This strategy not only reduces computational overhead but also removes the need for direct manipulation of the model's intermediate representations.

\subsection{Continuous Editing}

To endow editing models with continuous editing capabilities, several methods introduce trainable LoRA adapters to learn semantic directions mapped to attribute sliders~\cite{ConceptSliders,Alchemist,KontinuousKontext,SliderEdit}. These methods treat editing intensity as a controllable scalar, thereby achieving continuous manipulation effects. Furthermore, some studies train encoders to perform fine-grained manipulation at the token level within the text embedding space~\cite{SAEdit,CompassControl,ControllableColor,TextSlider}, enabling smooth control over editing attributes. However, these methods incur additional computational costs for training and rely on curated datasets, which significantly hampers their practical utility.

In addition, some training-free methods attempt to identify editing feature directions within the semantic latent space~\cite{ContinuousControl} and perform interpolation therein. However, features in the semantic space are not consistently continuous, making it challenging to identify a feature direction that balances editing accuracy with continuity. Other methods generate continuous editing effects by performing frame interpolation between the source image and the fully edited image~\cite{wan,frameWise,Sparsectrl}, or by interpolating within the diffusion feature space~\cite{FreeMorph,StableMorph}. Nonetheless, the generated intermediate states often suffer from abrupt transitions or exhibit artifacts and blurring~\cite{KontinuousKontext}.

VeloEdit is positioned as a training-free method. Distinct from the aforementioned strategies, we neither interpolate within the potentially discontinuous semantic space nor require the pre-generation of fully edited images for frame interpolation. Instead, we leverage the preservation velocity to intervene in the editing velocity within the editing regions, utilizing the robust generative and denoising capabilities of editing models~\cite{FluxKontext,QwenImage} to directly synthesize images with varying editing intensities.

\section{Method}
\label{sec:method}

\begin{figure}[tbp]
    \centering 
    \includegraphics[width=0.98\textwidth]{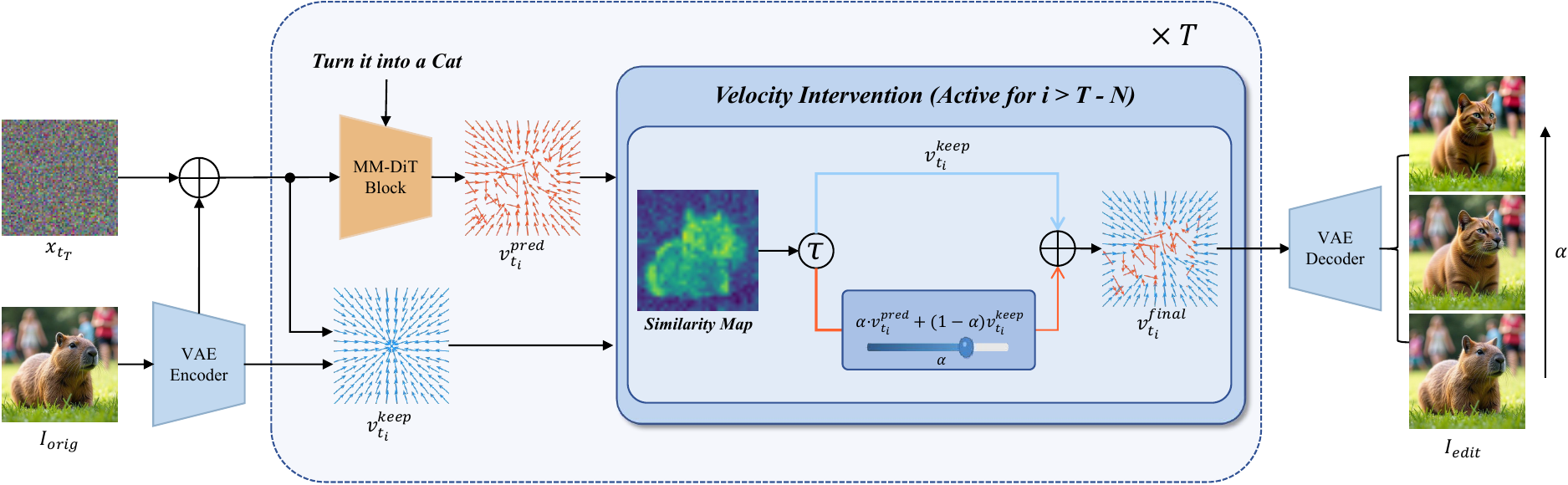}
    \caption{Overview of the proposed pipeline. We derive a spatial mask by analyzing the velocity discrepancy between preservation and editing flows. Our method explicitly preserves high similarity regions while blending low similarity regions, thereby yielding a sequence of edited results with smooth semantic transitions.} 
    \label{fig:pipeline} 
\end{figure}

\subsection{Preliminary}
\label{sec:preliminary}

\subsubsection{Flow Matching.} Flow matching~\cite{Flow,RectifiedFlow} introduces a generative method based on Continuous Normalizing Flows (CNFs), aiming to learn a deterministic transformation between a source distribution (e.g., noise) and a target distribution (data). The evolution of the probability density path \( p_t(\cdot) \) is governed by an Ordinary Differential Equation (ODE) parameterized by a time-dependent vector field \( v_t(x_t) \), such that the generative process is described by:

\[
\label{eq:1}
\frac{dx_t}{dt} = v_t(x_t).
\tag{1}
\]

To simplify vector field learning, rectified flow~\cite{Flow,RectifiedFlow} adopts straight line trajectories as a surrogate for more general transport paths, providing an optimal-transport–inspired formulation with reduced complexity. Given a data sample \( x_0 \sim P_{\text{data}} \) and a noise sample \( x_1 \sim \mathcal N(0, I) \), rectified flow defines the intermediate state \( x_t \) via linear interpolation:

\[\label{eq:2}
x_t = (1-t)x_0 + t x_1.
\tag{2}
\]

Under this construction, the conditional velocity field induced by the linear interpolation remains constant along the trajectory and is given by:

\[\label{eq:3}
u_t(x_t \mid x_0, x_1) = x_1 - x_0.
\tag{3}
\]

Accordingly, the objective of flow matching in this setting is to train a neural network \( v_\theta(x_t, t) \) to regress this conditional velocity field by minimizing:

\[ \label{eq:4}
\mathbb E_{t \sim \mathcal U(0,1)}
\mathbb E_{x_0 \sim P_{\text{data}}, x_1 \sim \mathcal N(0, I)}
\Vert v_\theta(x_t, t) - (x_1 - x_0) \Vert_2^2 .
\tag{4}
\]

By minimizing \cref{eq:4}, the model learns a velocity field that is consistent with the straight line coupling between noise and data distributions. This design encourages low curvature transport paths and enables efficient and high quality sampling with a small number of integration steps. As a result, rectified flow has been widely adopted in recent text-to-image generation and editing models~\cite{SD3,FluxKontext,QwenImage}.

\begin{algorithm}[tbp]
    \caption{VeloEdit}
    \label{alg:veloedit}
    \SetKwInOut{Input}{Input}
    \SetKwInOut{Output}{Output}

    \Input{Original image $I_{orig}$, edit prompt $P$, sampling steps $T$, intervention steps $N$, intervention threshold $\tau$, mixing weight $\alpha$}
    \Output{Edited image $I_{\text{edit}}$}
    Init: $x_1 \sim \mathcal{N}(0,I)$, $x_{orig} \leftarrow \text{Encoder}(I_{orig})$
    \BlankLine
    \For{$i \leftarrow T$ \KwTo $1$}{
        $t_i \leftarrow i/T,\ $ $t_{i-1} \leftarrow (i-1)/T$\;
        $v_{t_i}^{keep} \leftarrow (x_{t_i} - x_{orig}) / t_i,\ $ $v_{t_i}^{pred} \leftarrow \text{Model}(x_{t_i}, t_i, P)$\;
        $v_{t_i}^{final} \leftarrow v_{t_i}^{pred}$\;
        \If{$i > T - N$}{
            $S_{t_i} \leftarrow \frac{|v_{t_i}^{keep}|}{|v_{t_i}^{keep}| + |v_{t_i}^{keep} - v_{t_i}^{pred}|}$\;
            $M_{t_i}^{high} \leftarrow \mathbb{I}(S_{t_i} \ge \tau), \ $ $M_{t_i}^{low} \leftarrow \mathbb{I}(S_{t_i} < \tau)$\;
            $v_{t_i}^{final}[M_{t_i}^{high}] \leftarrow v_{t_i}^{keep}[M_{t_i}^{high}]$\;
            $v_{t_i}^{final}[M_{t_i}^{low}] \leftarrow (1-\alpha) \cdot v_{t_i}^{keep}[M_{t_i}^{low}] + \alpha \cdot v_{t_i}^{pred}[M_{t_i}^{low}]$\;
        }
        $x_{{t_{i-1}}} \leftarrow \text{Step}(x_{t_i}, v_{t_i}^{final})$\;
    }
    $I_{\text{edit}} \leftarrow \text{Decoder}(x_0)$\;
    \Return{$I_{\text{edit}}$}
\end{algorithm}

\subsection{Velocity Field Decomposition}
\label{sec:decomposition}

During inference, the predicted velocity $v_{t}^{pred}$ is entangled with both content preservation and editing guidance. To disentangle these factors, we decompose the velocity field. We define $v_t^{keep}$as the reference velocity required to reconstruct the source latent $x_{orig}$ from the current state $ x_t $. Under a linear flow assumption, it is formulated as:

\[
\label{eq:5}
v_{t}^{keep} = \frac{x_t - x_{orig}}{t}.
\tag{5}
\]

Consequently, the effective editing velocity $v_{t}^{diff}$, which governs the content modification, is defined as:

\[
\label{eq:6}
v_{t}^{diff} = v_{t}^{pred}-v_{t}^{keep}. \tag{6}
\]

Given that editing typically affects only local regions while leaving the background invariant, we expect $v_{t}^{pred}$ and $v_{t}^{keep}$ to deviate in edited areas but align closely in preserved regions. To quantify this spatial consistency, we introduce an element-wise similarity metric $S_t$. Formally, at coordinate $(i,j,k)$, the similarity $S_t^{(i,j,k)}$ is defined as:

\[\label{eq:7}
S_t^{(i,j,k)} = \frac{|v_{t}^{keep(i,j,k)}|}{|v_{t}^{keep(i,j,k)}| + |v_{t}^{diff(i,j,k)}| }.
\tag{7}
\]

For notational brevity, we omit the small constant $\epsilon$ typically added to the denominator for numerical stability. This metric is constrained to the range $(0, 1]$. Values approaching 1 ($S_t \to 1$) indicate strong alignment with the reference velocity, identifying background preservation regions. Conversely, values near 0 ($S_t \to 0$) imply significant deviation, characterizing the editing regions.

\subsection{Consistent Editing}
\label{sec:consistent}

Given a similarity threshold $\tau \in [0, 1]$, we define the velocity replacement strategy as follows:

\[\label{eq:8}
v_t^{replaced} = \begin{cases}
v_t^{keep}, & \text{if } S_t \geq \tau \\
v_t^{pred}, & \text{else} 
\end{cases}.
\tag{8}
\]

For notational brevity, we omit the spatial indices $(i,j,k)$ in the following sections. In high similarity regions ($S_t \geq \tau$), we enforce fidelity to the source image by substituting the prediction with $v_t^{keep}$. Conversely, in low similarity regions, we retain $v_t^{pred}$ to enable content modification. To avoid over constraining the generative process, we apply this intervention only during the initial $N$ steps of the $T$-step denoising trajectory:

\[\label{eq:9}
v_t^{final} = 
\begin{cases}
    v_t^{replaced}, & \text{if } t > 1 - N/T \\
    v_t^{pred}, & \text{else}
\end{cases}.
\tag{9}
\]

By anchoring non-edited regions to the source content, our selective replacement mechanism ensures structural invariance while affording the model the necessary flexibility to execute local edits.

\subsection{Continuous Editing}
\label{sec:continuous}

For low similarity regions ($S_t < \tau$), we introduce a velocity blending strategy to enable continuous modulation of the editing intensity:

\[\label{eq:10}
v_{t}^{blend} = (1 - \alpha)  \cdot v_{t}^{keep} + \alpha \cdot v_{t}^{pred}, 
\tag{10}
\]

where $\alpha \in \mathbb{R}$ serves as the blending coefficient. Specifically, when $\alpha \in [0,1]$, the editing effect smoothly interpolates between the source image and the fully edited output. Conversely, values outside this range ($\alpha < 0$ or $\alpha > 1$) result in the extrapolation of the editing effect. By integrating selective replacement with velocity blending, the unified intervention formulation is defined as:

\[\label{eq:11}
v_{t}^{final} = v_{t}^{keep} \cdot \mathbb{I}(S_t \geq \tau) + v_{t}^{blend} \cdot \mathbb{I}(S_t < \tau),  \tag{11}
\]

where $\mathbb{I}(\cdot)$ denotes the indicator function, which takes the value of 1 if the condition in the parentheses is satisfied, and 0 otherwise. The complete pipeline of VeloEdit is outlined in \cref{fig:pipeline} and \cref{alg:veloedit}.

\begin{figure}[btp]
    \centering 
    \includegraphics[width=0.85\textwidth]{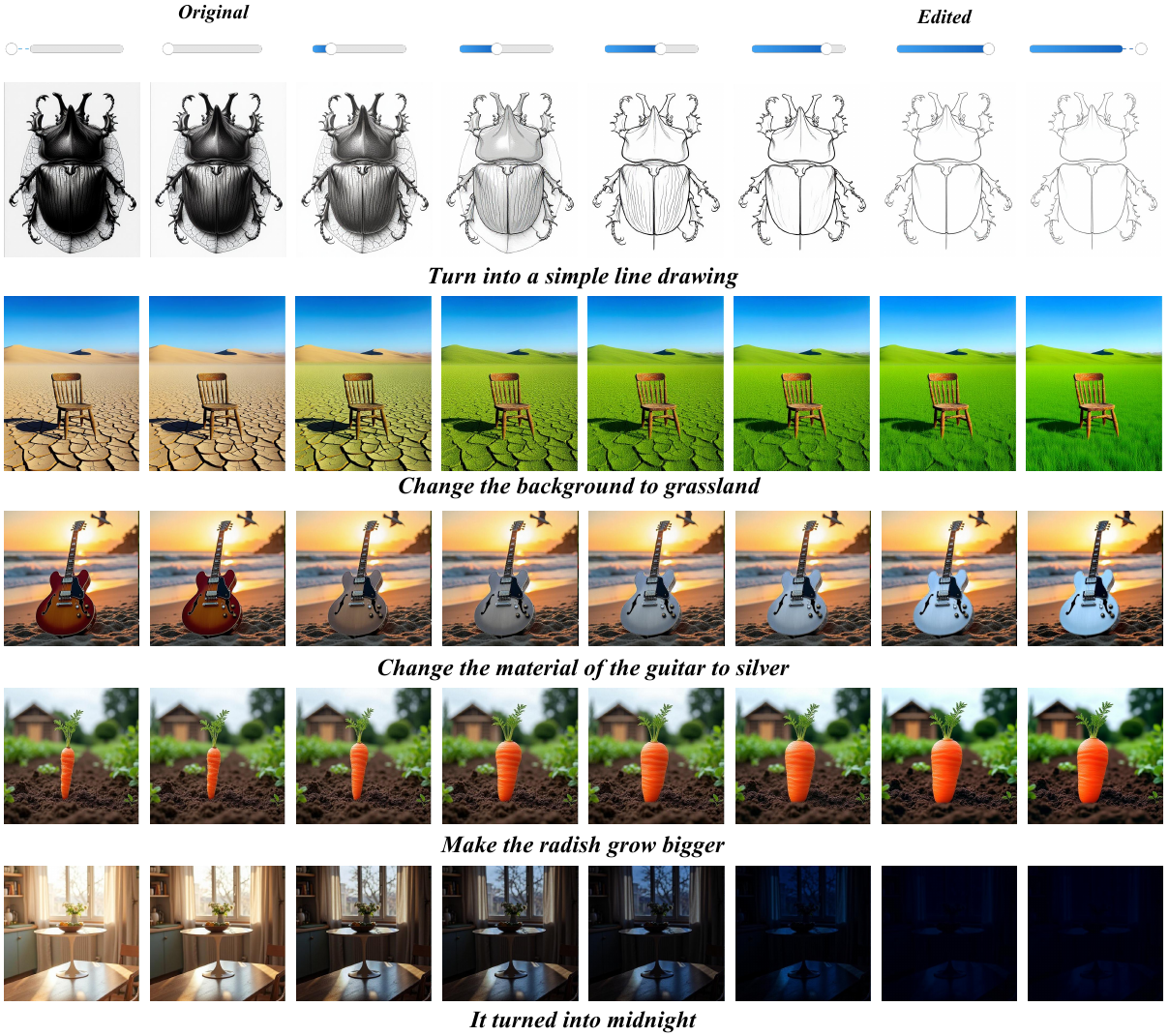}
    \caption{Visual editing results on GPT-Image-Edit~\cite{GptImage} and Subject200K~\cite{Ominicontrol}. VeloEdit achieves smooth control over both local and global editing intensities.} 
    \label{fig:exp_img_all} 
\end{figure}

\section{Experiments}
\label{sec:experiments}

We conduct a comprehensive quantitative and qualitative evaluation of VeloEdit on Flux.1 Kontext and Qwen-Image-Edit. The results show that VeloEdit improves editing consistency and enables continuous editing, demonstrating its generality. In addition, we compare VeloEdit with various baselines for consistency preservation and continuous editing, and show that our method achieves competitive performance in maintaining consistency while providing smoother and more precise control over continuous editing. For more experimental results and explanations, please refer to the supplementary material.

\subsection{Details}
\label{sec:details}

In this section, we present the implementation details of our experiments.

\textbf{Benchmark.} We conduct a comprehensive evaluation against baselines on PIEbench, which comprises 700 image-instruction pairs covering diverse tasks such as object modification, addition/removal, and changes in pose, color, material, and background. While tasks like object addition/removal inherently lack continuous transition semantics, we include them to rigorously analyze the failure modes and boundary conditions of the methods. Furthermore, to assess cross-dataset generalization of VeloEdit, we extend our evaluation to the Subject200K~\cite{Ominicontrol} and GPT-Image-Edit~\cite{GptImage} datasets.

\textbf{Settings.} We implement VeloEdit on top of Flux.1 Kontext and Qwen-Image-Edit. For consistency experiments, we adopt the default configuration: intervention threshold $\tau=0.4$, sampling steps $T=6$, and intervention steps $N=1$. For continuity evaluations, we adjust $\tau=0.8$ and assess all methods at five uniformly spaced edit-strength levels within the range $[0.2, 1]$. All experiments were conducted on NVIDIA H800 GPU (80GB).

\textbf{Metrics:} In the consistency comparison experiments, we use PSNR~\cite{PSNR} and SSIM~\cite{SSIM} to quantify background preservation, and CLIP similarity (CLIP-Sim.)~\cite{clipSimilar} to evaluate the editing effect. In the continuity comparison experiments, we evaluate methods based on continuity, instruction adherence, and consistency preservation. Specifically, following the protocol in~\cite{KontinuousKontext}, we employ the triangular defect $\delta_{smooth}$ to measure the smoothness of editing results, utilizing Dream-Sim~\cite{Dreamsim} as the distance metric. Instruction adherence is evaluated via CLIP directional similarity (CLIP-Dir.)~\cite{clipdir}, averaged across all editing strengths. For consistency preservation, we compute the $L1$ and $L2$ distances specifically within the non-edited regions defined by PIEbench masks.

\subsection{Main Results}
\label{sec:main_results}

\subsubsection{Qualitative Results}
\label{sec:qualitative}

\begin{table*}[ht]
\centering
\caption{Consistency evaluation results on PIEbench. Our method achieves competitive performance compared to methods based on inversion and attention map injection. Baseline results are cited from~\cite{ProEdit}, and the best and second-best results are indicated in bold and underlined, respectively.}
\label{tab:compare_similar}
\begin{tabular}{lccc}
\toprule
{Method} & \quad PSNR$\uparrow$ \quad & \quad SSIM ($\times 10^2$)$ \uparrow$ \quad &  \quad {CLIP-Sim.$\uparrow$} \quad \\
\midrule
P2P~\cite{ptp} & 17.87 & 71.14 & 25.01 \\
PnP~\cite{pnp} & 22.28 & 79.05 & 25.41 \\
MasaCtrl~\cite{Masactrl} & 22.17 & 79.67 & 23.96 \\
RF-Inversion~\cite{RFInver} & 20.82 & 71.92 & 25.20 \\
RF-Solver~\cite{RFSolver} & 22.90 & 81.90 & 26.00 \\
UniEdit~\cite{Uniedit} & \underline{24.10} & \textbf{84.86} & 26.97 \\
\arrayrulecolor{gray!60} 
\cmidrule(lr){1-4}
\arrayrulecolor{black}
Flux.1 Kontext~\cite{FluxKontext} & 16.44 & 64.70 & \textbf{30.65} \\
\quad + VeloEdit & \textbf{28.10} & \underline{82.20} & \underline{30.49} \\
\bottomrule
\end{tabular}
\end{table*}

\begin{figure}[btp] 
    \centering 
    \includegraphics[width=1\textwidth]{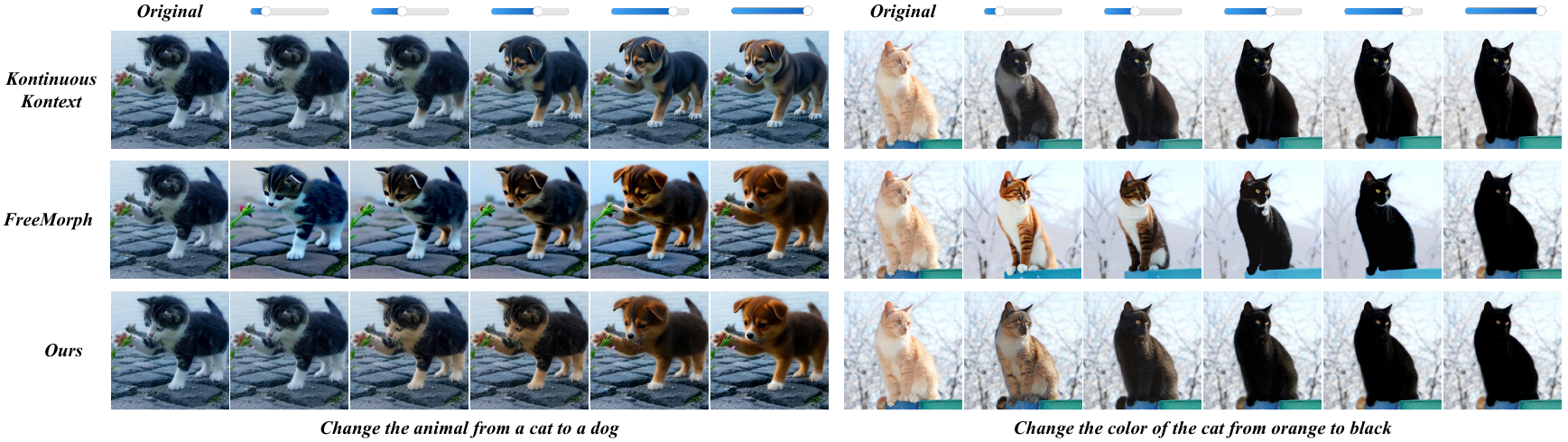}
    \caption{Qualitative comparison results. VeloEdit generates edited results with superior consistency and continuity, effectively avoiding background detail alteration and drift.} 
    \label{fig:compare_1} 
\end{figure}

\begin{figure}[btp] 
    \centering 
    \includegraphics[width=1\textwidth]{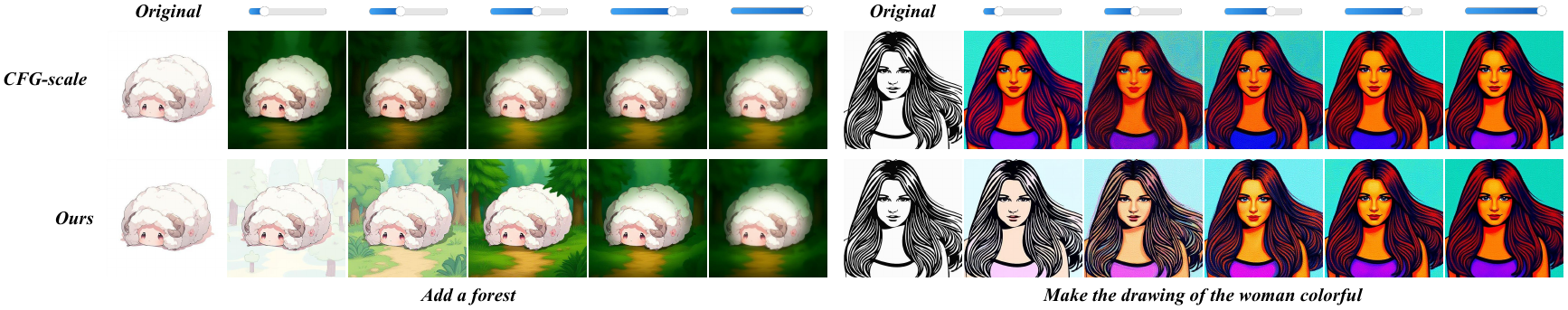}
    \caption{Qualitative comparison with CFG-scale guidance. VeloEdit produces consistent and continuous edits, while CFG-scale fails to maintain continuity.} 
    \label{fig:compare_2} 
\end{figure}

\begin{figure}[btp]
    \centering 
    \includegraphics[width=1\textwidth]{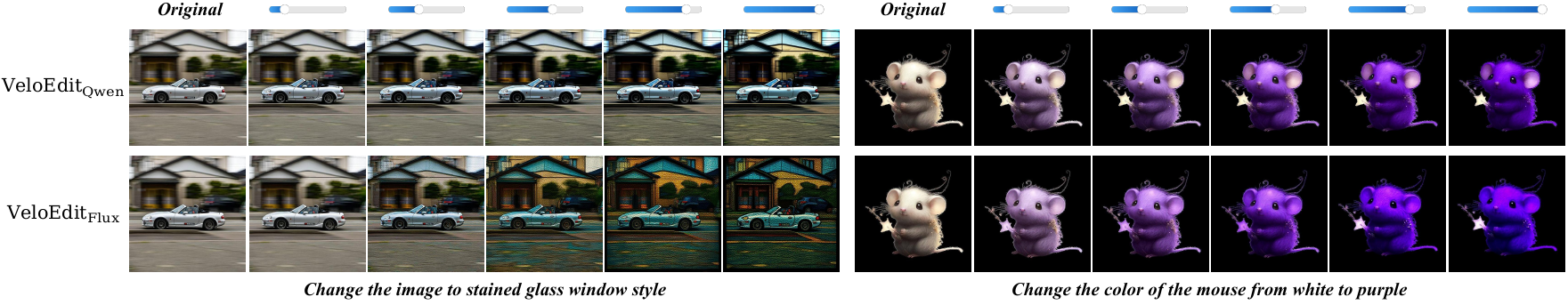}
    \caption{Qualitative comparison results. Our method is effective in both Flux.1 Kontext~\cite{FluxKontext} and Qwen-Image-Edit~\cite{QwenImage}, and can generate continuous and smooth editing results.} 
    \label{fig:vis_flux_qwen} 
\end{figure}

We qualitatively evaluate VeloEdit to demonstrate its versatility across diverse editing tasks and model architectures. As illustrated in \cref{fig:exp_img_all}, VeloEdit generates smooth, continuous editing trajectories that enable fine-grained control over editing intensity. Our method effectively handles a wide array of scenarios, ranging from global adjustments (e.g., style transfer, background modification, and colorization) to local manipulations (e.g., object replacement and attribute modification). Comparisons against existing training-based and training-free baselines in \cref{fig:compare_1} highlight that VeloEdit yields superior continuity while mitigating background drift. Furthermore, \cref{fig:compare_2} contrasts our approach with CFG guidance strategies; the results reveal that naively scaling guidance magnitude fails to produce smooth transitions. Finally, results on Flux.1 Kontext and Qwen-Image-Edit (\cref{fig:vis_flux_qwen}) consistently exhibit smooth editing progressions, demonstrating VeloEdit's robust cross-model generalizability.

\subsubsection{Quantitative Results}
\label{sec:quanlitative}

\begin{table}[htbp]
\centering
\caption{Quantitative experiments on PIEbench. VeloEdit achieves the best or second-best metrics compared with training-free and training-based methods.
}
\label{tab:compare_all}
\begin{tabular}{lcccc}
\hline
Method & $\delta_{\text{smooth}} \downarrow$ & CLIP-Dir. $\uparrow$ &  $L_1 \downarrow$ & $L_2 \downarrow$ \\
\hline
\textbf{Training-Free}\\
Freemorph~\cite{FreeMorph} & 0.354 & 0.147  & 0.142  & 0.211 \\
CFG-scale~\cite{FluxKontext,ClassifierFree} & 3.362 & \textbf{0.379}  & 0.140  & 0.209 \\
\arrayrulecolor{gray!60} 
\cmidrule(lr){1-5}      
\arrayrulecolor{black}
\textbf{Training-Based}\\
KontinuousKontext~\cite{KontinuousKontext} & 0.280 & \underline{0.219} & \underline{0.083}  & \underline{0.132} \\
\arrayrulecolor{gray!60} 
\cmidrule(lr){1-5}      
\arrayrulecolor{black}
VeloEdit & \textbf{0.246} & \underline{0.294}  & \textbf{0.074}  & \textbf{0.116} \\
\hline
\end{tabular}
\end{table}

We first conduct a quantitative evaluation on PIEbench~\cite{PIEbench}, comparing VeloEdit against multiple feature injection-based consistency preservation baselines. As shown in \cref{tab:compare_similar}, compared to feature injection methods, our approach achieves the best PSNR and second-best SSIM metrics. Notably, it significantly improves consistency preservation in non-edited regions with only a minimal loss in CLIP-Sim. score.

Subsequently, we quantitatively evaluate VeloEdit against multiple continuous editing baselines on PIEbench. We assess continuity, instruction following, and consistency to analyze the smoothness and content preservation of the generated editing trajectories. Specifically, we employ Flux.1 Kontext~\cite{FluxKontext} to generate fully edited target images and leverage Freemorph~\cite{FreeMorph} to synthesize intermediate editing states. We compare our method against Freemorph, KontinuousKontext~\cite{KontinuousKontext} (the current state-of-the-art open-source continuous editing model) as well as CFG guidance strategies across varying strengths. As shown in \cref{tab:compare_all}, our method achieves the best or second-best performance across all metrics, demonstrating the effectiveness of VeloEdit. Notably, while CFG-scale yields the maximum CLIP-Dir. scores, it suffers from a significant degradation in $\delta_\text{smooth}$. This trade-off is further elucidated in \cref{fig:compare_2}, where CFG-scale prematurely produces high-intensity editing effects.

To evaluate the cross-model generalization capability of VeloEdit, we apply VeloEdit to Qwen-Image-Edit~\cite{QwenImage} using the identical hyperparameter configuration as Flux.1 Kontext, and test its performance on PIEbench, with the corresponding results presented in \cref{tab:flux_qwen}. The experimental results demonstrate that VeloEdit can be integrated into various editing models and endow them with consistent and continuous editing capabilities. 

\begin{table}[tbp]
\centering
\caption{Cross-model generalization performance on the PIEbench. VeloEdit demonstrates consistent effectiveness when integrated with Flux.1 Kontext and Qwen-Image-Edit.}
\label{tab:flux_qwen}
\begin{tabular}{lcccc}
\hline
Method & $\delta_{\text{smooth}} \downarrow$ & CLIP-Dir. $\uparrow$ &  $L_1 \downarrow$ & $L_2 \downarrow$ \\
\hline
$\text{VeloEdit}_{\text{Flux.1 Kontext}}$ & 0.246 & 0.294  & \textbf{0.074}  & 0.116 \\
\arrayrulecolor{gray!60} 
\cmidrule(lr){1-4}
\arrayrulecolor{black}
$\text{VeloEdit}_{\text{Qwen-Image-Edit}}$ & \textbf{0.207} & \textbf{0.349}  & 0.075  & \textbf{0.104} \\
\hline
\end{tabular}
\end{table}

\begin{table}[tbp]
\centering
\caption{Ablation study on $\tau$. Performance of VeloEdit using different $\tau$ values across Flux.1 Kontext and Qwen-Image-Edit.}
\label{tab:abs_tau}
\small
\begin{tabular}{l|cccc|cccc}
\hline
\multirow{2}{*}{$\tau$} & \multicolumn{4}{c|}{Flux.1 Kontext} & \multicolumn{4}{c}{Qwen-Image-Edit} \\
\cline{2-9}
 & $\delta_{\text{smooth}} \downarrow$ & CLIP-Dir. $\uparrow$ & $\quad L_1 \downarrow$ \quad  & $\quad L_2 \downarrow \quad $ & $\delta_{\text{smooth}} \downarrow$ & CLIP-Dir. $\uparrow$ & $\quad L_1  \downarrow \quad$  & $\quad L_2 \downarrow \quad$  \\
\hline
0.0 & 0.038 & 0.079 & \textbf{0.034} & \textbf{0.052} & 0.024 & 0.149 & \textbf{0.023} &\textbf{ 0.038} \\
0.2 & 0.751 & 0.168 & 0.041 & 0.063 & 0.921 & 0.195 & 0.025 & 0.040 \\
0.4 & 0.439 & 0.249 & 0.055 & 0.088 & 0.432 & 0.279 & 0.036 & 0.056 \\
0.6 & 0.303 & 0.285 & 0.067 & 0.107 & 0.242 & 0.334 & 0.058 & 0.084 \\
0.8 & \textbf{0.246} & 0.294 & 0.074 & 0.116 & 0.207 & 0.349 & 0.075 & 0.104 \\
1.0 & 0.255 & \textbf{0.296} & 0.075 & 0.117 & \textbf{0.195} & \textbf{0.351} & 0.078 & 0.108 \\
\hline
\end{tabular}
\end{table}

\begin{table}[tbp]
\centering
\caption{Ablation study on $N$. Performance of VeloEdit using different $N$ values across Flux.1 Kontext and Qwen-Image-Edit.}
\label{tab:abs_N}
\small
\begin{tabular}{l|cccc|cccc}
\hline
\multirow{2}{*}{$N$} & \multicolumn{4}{c|}{Flux.1 Kontext} & \multicolumn{4}{c}{Qwen-Image-Edit} \\
\cline{2-9}
 & $\delta_{\text{smooth}}\downarrow$ & CLIP-Dir.$\uparrow$ & $\quad L_1\downarrow$ \quad  & $\quad L_2\downarrow \quad $ & $\delta_{\text{smooth}}\downarrow$ & CLIP-Dir.$\uparrow$ & $\quad L_1\downarrow \quad$  & $\quad L_2\downarrow \quad$  \\
\hline
1 & 0.246 & \textbf{0.294} & 0.074 & 0.116 & 0.207 & \textbf{0.351} & 0.078 & 0.108 \\
2 & 0.220 & 0.207 & 0.063 & 0.098 & \textbf{0.133} & 0.271 & 0.059 & 0.082 \\
3 & \textbf{0.206} & 0.167 & 0.057 & 0.089 & 0.153 & 0.215 & 0.048 & 0.068 \\
4 & 0.229 & 0.138 & 0.052 & 0.083 & 0.259 & 0.170 & 0.041 & 0.059 \\
5 & 0.290 & 0.116 & 0.048 & 0.077 & 0.331 & 0.106 & 0.034 & 0.050 \\
6 & 0.387 & 0.087 & \textbf{0.043} & \textbf{0.069} & 0.490 & 0.078 & \textbf{0.031} & \textbf{0.046} \\
\hline
\end{tabular}
\end{table}

\begin{figure}[hbtp]
    \centering 
    \includegraphics[width=0.9\textwidth]{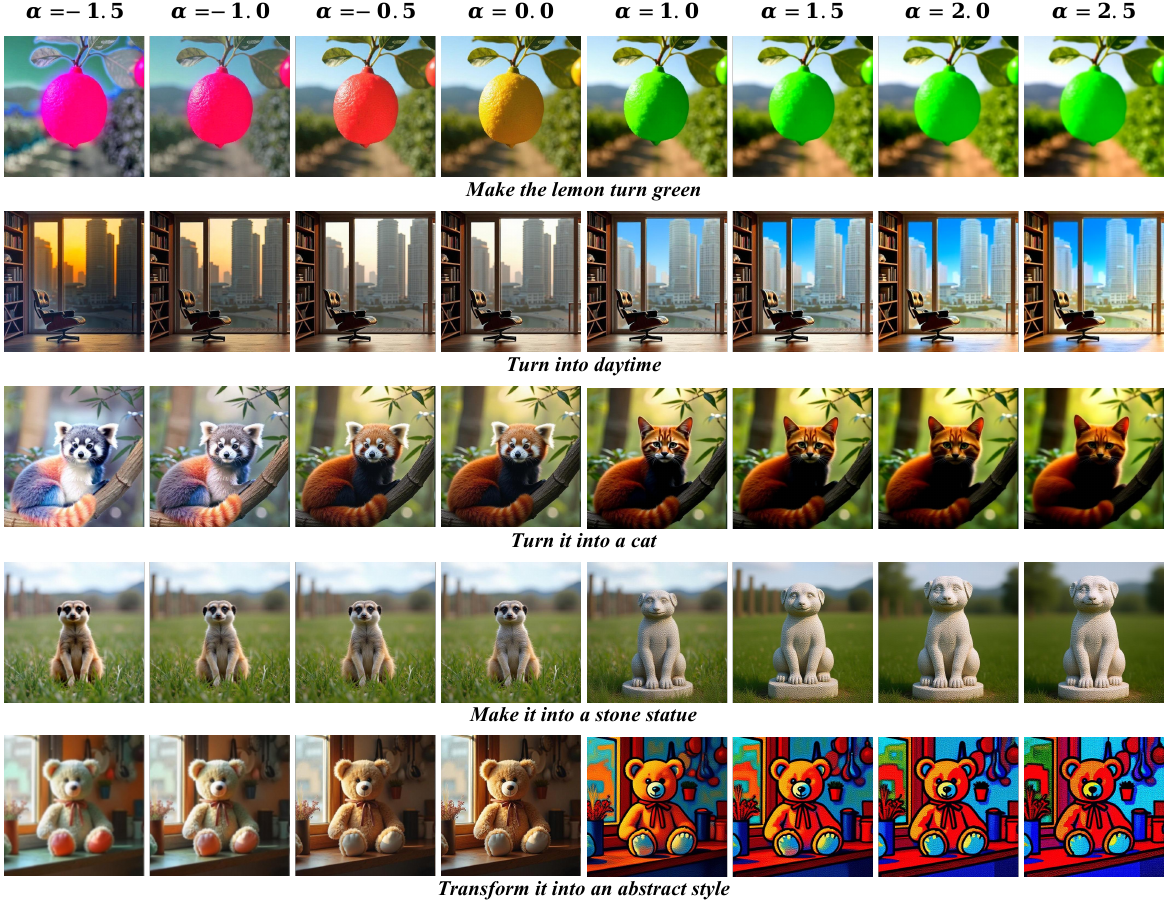}
    \caption{Visualizations of extrapolation for $\alpha$. By employing blending intensities beyond the standard $[0, 1]$ range, VeloEdit can achieve either inverse semantic effects ($\alpha < 0$) or intensified editing results ( $\alpha > 1$) relative to the prompt.} 
    \label{fig:abs_alpha} 
\end{figure}

\subsection{Ablation Studies}
\label{sec:ablation}

In this section, we conduct ablation studies on the hyperparameters $\tau$ and $N$ of VeloEdit, as reported in \cref{tab:abs_tau,tab:abs_N}. The results demonstrate that our method exhibits strong robustness regarding $\tau$, and performance degradation, such as loss of continuity or unexpected editing behaviors, occurs only at extreme values. Meanwhile, the ablation study on $N$ indicates that intervening during the initial one or two denoising steps strikes a favorable balance between continuity and instruction adherence. This finding further corroborates our observations in \cref{fig:f_1,fig:f_2}. Furthermore, although the value of $\alpha$ is theoretically unbounded, visualization results from \cref{fig:exp_img_all,fig:abs_alpha} suggest that constraining $\alpha$ within the range of $[-1.0, 2.0]$ yields superior visual quality.

\section{Conclusion}
\label{conclusion}

In this paper, we presented VeloEdit, a generic, training-free method designed to enhance consistency and enable continuous capabilities for instruction-based image editing models. By evaluating the disparity between preservation and editing velocities, VeloEdit effectively decomposes the velocity field. Specifically, during the early denoising stages, it substitutes the editing velocity with the preservation velocity in high similarity regions to preserve structural integrity, while employing a velocity blending strategy in low similarity regions for smooth intensity modulation. Experiments with Flux.1 Kontext and Qwen-Image-Edit demonstrate that our method significantly improves visual consistency and editing continuity, achieving these gains without modifying internal attention mechanisms or requiring additional training.

\clearpage

\printbibliography

\clearpage

\section{Supplementary Material}
\label{sec:supplementary}

\subsection{Metrics}
\label{sec:app_metrics}

\subsubsection{Consistency Metric }
\label{sec:app_consistency}

We employ PSNR and SSIM to evaluate consistency preservation in non-edited regions. Specifically, utilizing the mask information from PIEbench, we precisely separate the edited and non-edited regions, calculating consistency metrics exclusively within the non-edited areas. Higher scores indicate superior consistency preservation capabilities in these regions. Furthermore, we use CLIP similarity to assess the overall editing effect on the entire image, where a higher score signifies stronger instruction adherence.

\subsubsection{Continuity Metric}
\label{sec:app_continuity}

Following the protocol in KontinuousKontext, we employ $\delta_{\text{smooth}}$ to evaluate the smoothness of the editing trajectory. Given an original image $x$, we utilize an editing instruction $P$ along with uniformly sampled edit strengths $\{{\alpha_1}, {\alpha_2}, \dots, {\alpha_N}\}$ to generate a sequence of images at various edit strengths $\{x_{\alpha_1}, x_{\alpha_2}, \dots, x_{\alpha_N}\}$. We set $x_{\alpha_0} = x$ to obtain a sequence of $N+1$ images. Subsequently, we apply DreamSim as the distance metric $d(\cdot, \cdot)$ to compute the difference between any two images. We define $\delta_{\text{smooth}}$ as follows:

\[
\delta_\text{smooth} = \max_i \frac{d(x_{\alpha_i},x_{\alpha_{i+1}})+d(x_{\alpha_{i+1}},x_{\alpha_{i+2}})-d(x_{\alpha_i},x_{\alpha_{i+2}})}{d(x_{\alpha_i},x_{\alpha_{i+2}})},\  i = 0,..., N-2.
\]

Furthermore, we employ the CLIP directional similarity $\text{CLIP-Dir.}$ to evaluate instruction adherence. The calculation for $\text{CLIP-Dir.}$ is defined as:

\[
\text{CLIP-Dir.} = \frac{\sum_{i=1}^{N} (\text{CLIP-Sim.}(x_{\alpha_{i}},x) / \alpha_i)}{N}.
\]

Subsequently, to evaluate background consistency preservation during continuous editing, we utilize masks to separate the edited regions from the non-edited regions. We then employ the $L_1$ and $L_2$ metrics to measure the distance between the edited and original images exclusively within the non-edited areas. Finally, following the aggregation protocol of $\text{CLIP-Dir.}$, we accumulate the $L_1$ and $L_2$ distances across different edit strengths.

\subsection{Additional Qualitative Results}
\label{sec:app_addQuaRes}

\begin{figure}[btp]
    \centering 
    \includegraphics[width=0.85\textwidth]{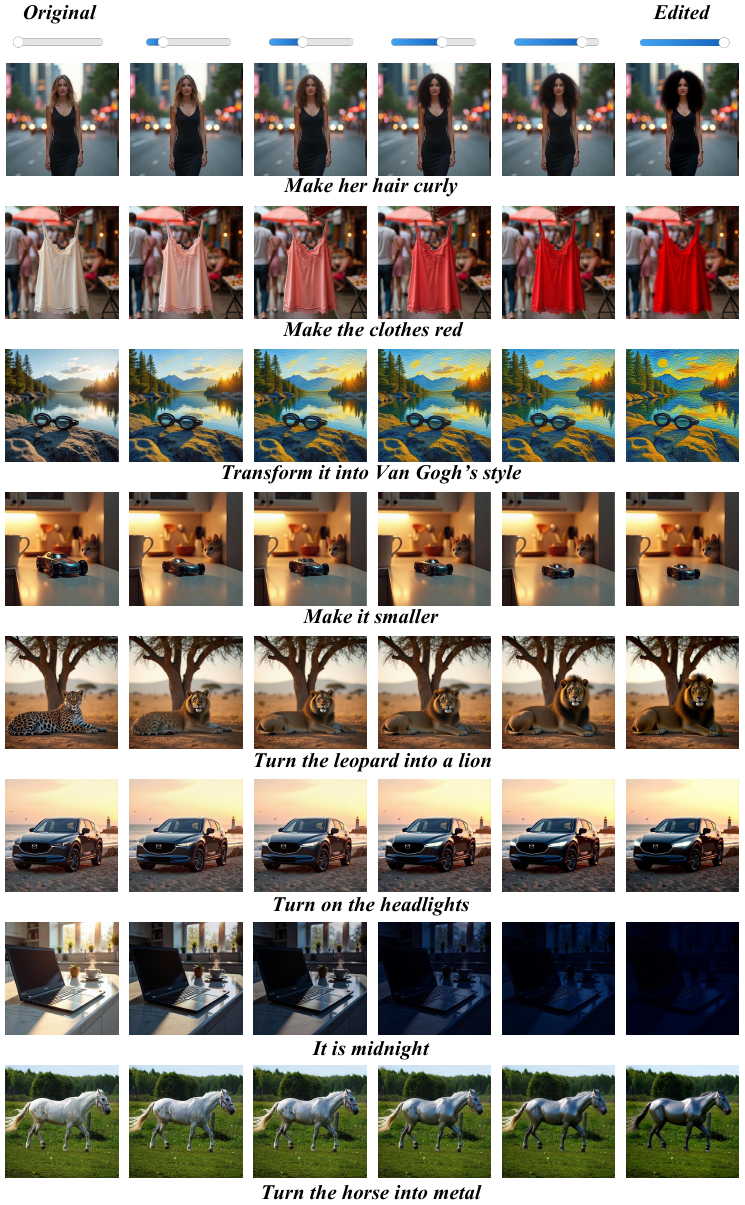}
    \caption{Additional visual results on Subject200K. VeloEdit generates consistent and continuous results across diverse editing tasks.} 
    \label{fig:add_f_1} 
\end{figure}

\begin{figure}[btp]
    \centering
    \includegraphics[width=0.9\textwidth]{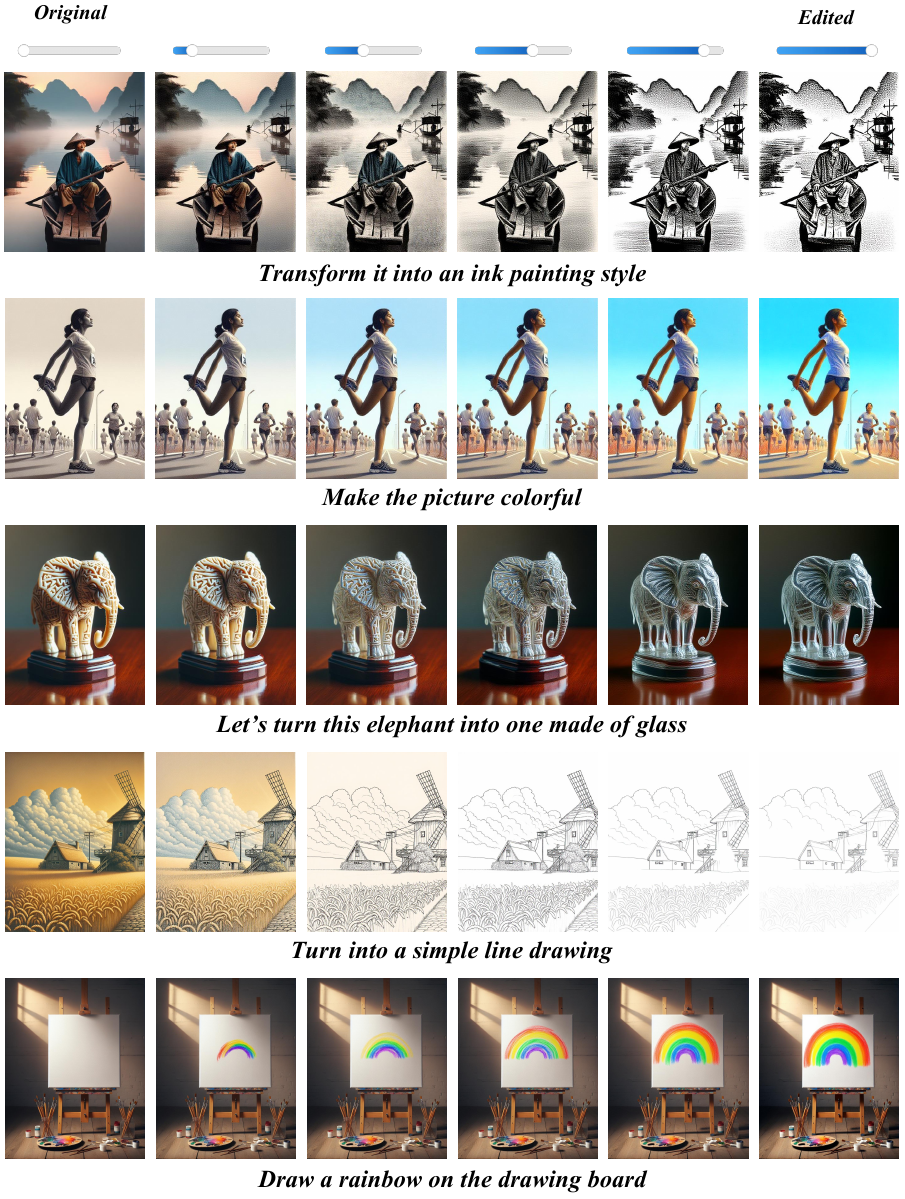}
    \caption{Additional visual results on GPT-Image-Edit. VeloEdit generates consistent and continuous results across diverse editing tasks.} 
    \label{fig:add_f_2} 
\end{figure}

\begin{figure}[btp]
    \centering
    \includegraphics[width=0.9\textwidth]{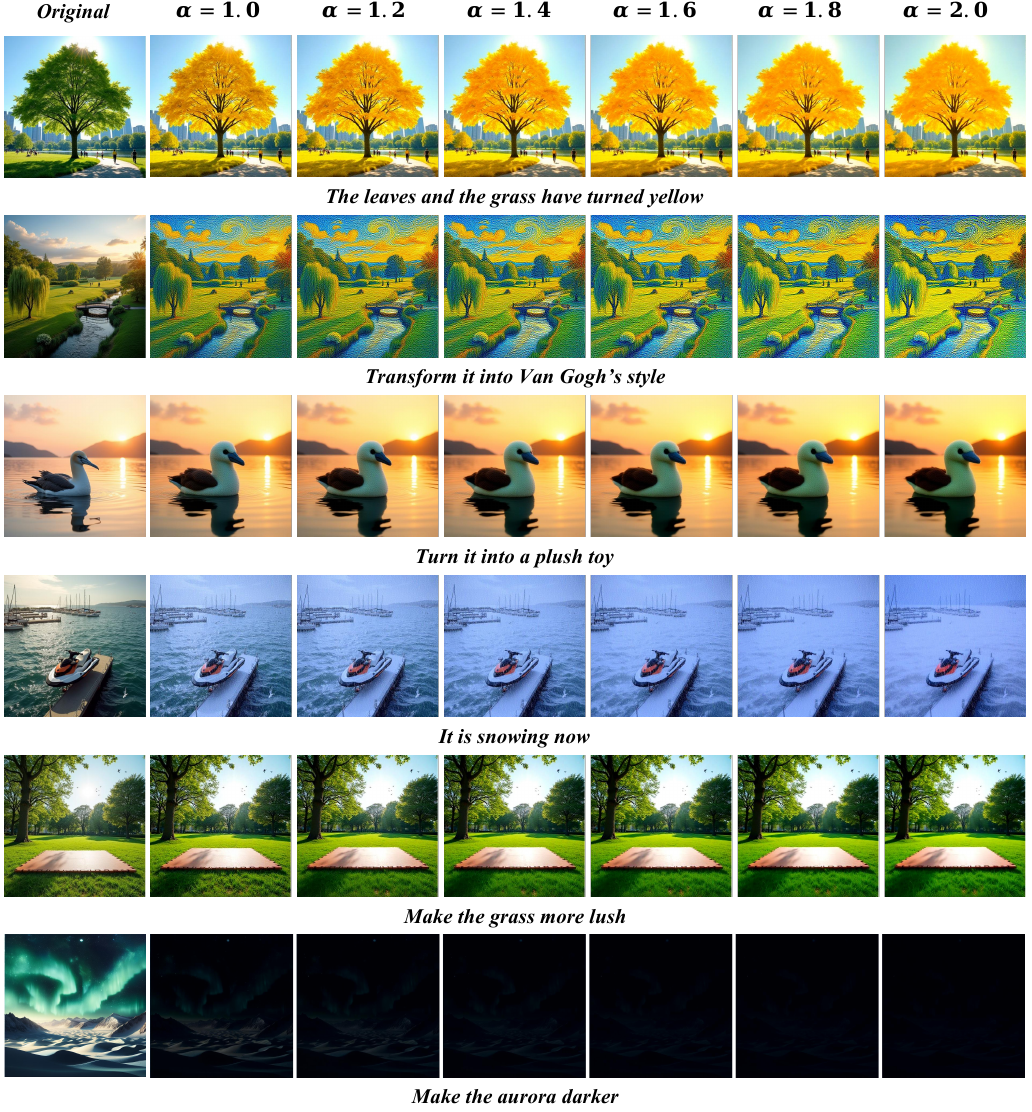}
    \caption{By employing a larger $\alpha$, VeloEdit expands the editing boundaries of the foundation model and achieves more pronounced continuous editing outcomes.} 
    \label{fig:add_f_3} 
\end{figure}

\begin{figure}[btp]
    \centering
    \includegraphics[width=0.98\textwidth]{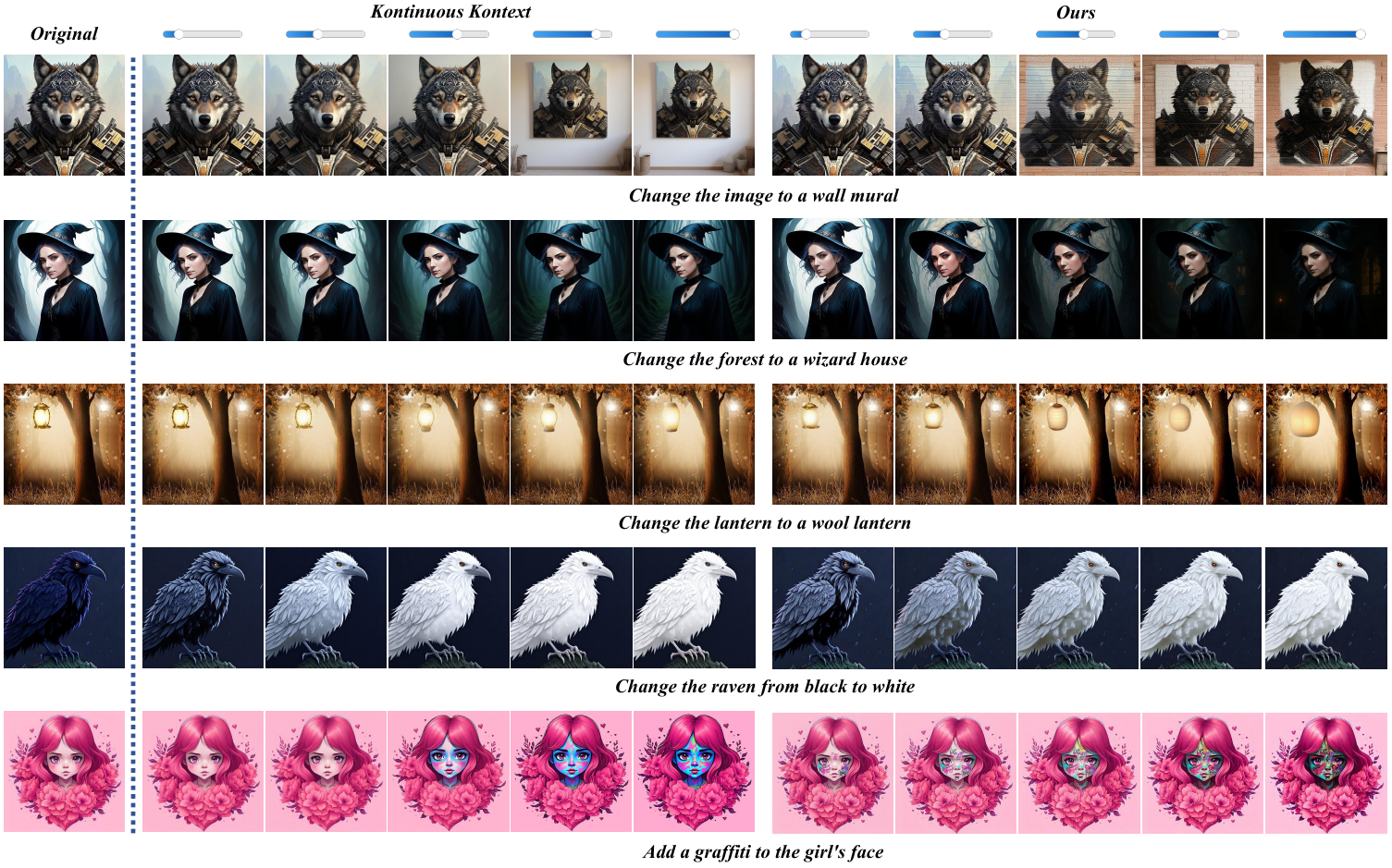}
    \caption{Additional qualitative comparison with Kontinuous Kontext. Our method better preserves consistency in unedited regions and generates more continuous editing results.} 
    \label{fig:add_f_4} 
\end{figure}

\begin{figure}[btp]
    \centering
    \includegraphics[width=0.99\textwidth]{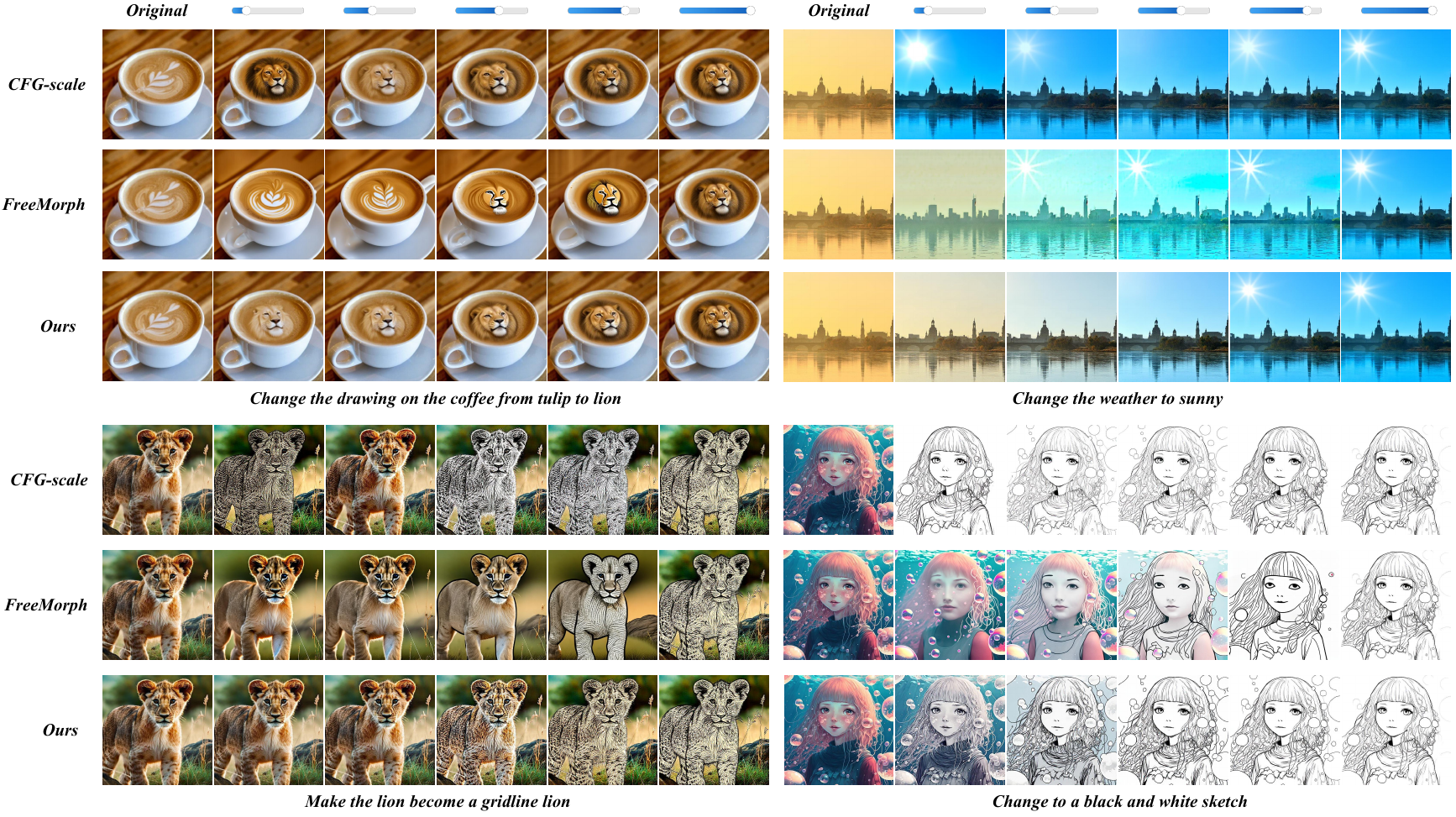}
    \caption{Additional qualitative comparison with FreeMorph and CFG-scale. Our method better preserves consistency in unedited regions and generates more continuous editing results.} 
    \label{fig:add_f_5} 
\end{figure}

\begin{figure}[btp]
    \centering
    \includegraphics[width=0.99\textwidth]{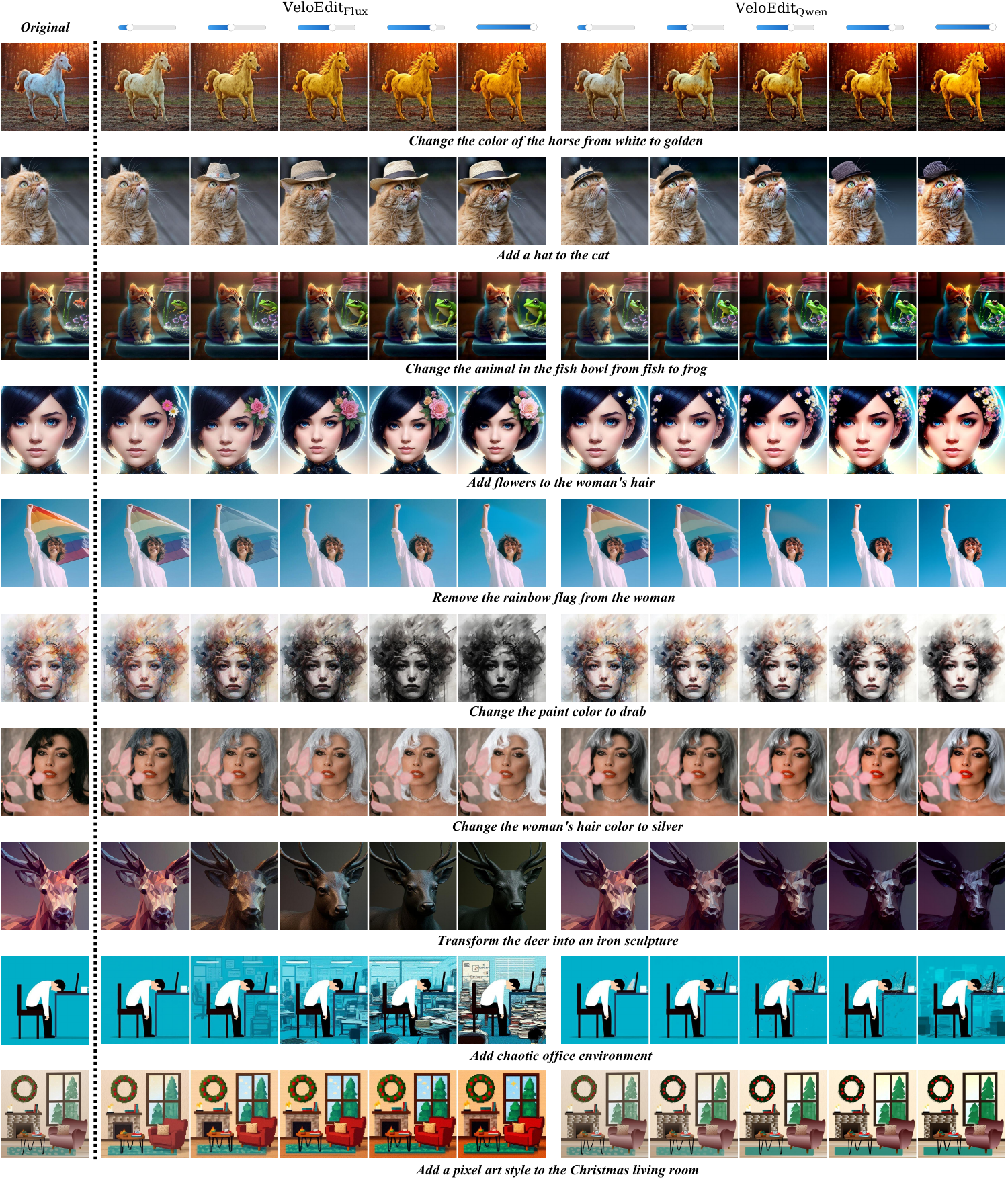}
    \caption{Additional visual results. Our method is effective in both Flux.1 Kontext and Qwen-Image-Edit, and can generate continuous and smooth editing results.} 
    \label{fig:add_f_6} 
\end{figure}

In this section, we present additional visual results. \Cref{fig:add_f_1,fig:add_f_2} demonstrate the effectiveness of our method across various tasks and image resolutions. \Cref{fig:add_f_3} illustrates its extrapolation capabilities. \Cref{fig:add_f_4,fig:add_f_5} provides visual comparisons with other methods, indicating that VeloEdit generates more consistent and continuous editing results. Finally, \Cref{fig:add_f_6} displays the editing outcomes when integrating our method with Flux.1 Kontext and Qwen-Image-Edit, demonstrating its strong generalization capabilities across different models.

\subsection{Additional Quantitative Results}
\label{sec:app_imp_addQuaRes}

We report the consistency and continuity metrics across various tasks on PIEBench in \cref{tab:benchmark_comparison}. The results demonstrate that VeloEdit achieves the best consistency and instruction adherence across all tasks, while attaining the best or second-best continuity in multiple tasks. Furthermore, we report the resource consumption introduced by VeloEdit in \cref{tab:time_compare}, and the results verify that our method introduces almost no additional time cost, with an extra time consumption of less than $0.02\%$.

\begin{table*}[tbp]
\centering
\caption{Performance comparison of VeloEdit and other continuous editing baselines across different editing tasks.}
\label{tab:benchmark_comparison}
\resizebox{0.9\textwidth}{!}{%
\begin{tabular}{llcccc}
\toprule
\textbf{Edit Type} & Method & $\delta_{\text{smooth}} \downarrow$ & CLIP-Dir. $\uparrow$ &  $L_1 \downarrow$ & $L_2 \downarrow$ \\
\midrule
\multirow{3}{*}{Random}
 & FreeMorph & 0.302 & 0.177 & 0.150 & 0.223 \\
 & KontinuousKontext & \textbf{0.119} & 0.251 & \textbf{0.089} & 0.136 \\
 & VeloEdit & 0.151 & \textbf{0.321} & \textbf{0.089} & \textbf{0.135} \\
\midrule
\multirow{3}{*}{Change Object}
 & FreeMorph & 0.410 & 0.208 & 0.159 & 0.253 \\
 & KontinuousKontext & 0.379 & 0.236 & 0.091 & 0.155 \\
 & VeloEdit & \textbf{0.277} & \textbf{0.384} & \textbf{0.086} & \textbf{0.151} \\
\midrule
\multirow{3}{*}{Add Object}
 & FreeMorph & \textbf{0.394} & 0.096 & 0.111 & 0.174 \\
 & KontinuousKontext & 0.416 & 0.158 & 0.066 & 0.108 \\
 & VeloEdit & 0.439 & \textbf{0.362} & \textbf{0.054} & \textbf{0.089} \\
\midrule
\multirow{3}{*}{Delete Object}
 & FreeMorph & 0.363 & 0.186 & 0.147 & 0.233 \\
 & KontinuousKontext & 0.661 & 0.302 & 0.083 & 0.138 \\
 & VeloEdit & \textbf{0.347} & \textbf{0.311} & \textbf{0.057} & \textbf{0.092} \\
\midrule
\multirow{3}{*}{Change Attr. Content}
 & FreeMorph & 0.357 & 0.066 & 0.142 & 0.215 \\
 & KontinuousKontext & \textbf{0.170} & 0.146 & \textbf{0.075} & 0.123 \\
 & VeloEdit & 0.376 & \textbf{0.151} & 0.077 & \textbf{0.114} \\
\midrule
\multirow{3}{*}{Change Attr. Pose}
 & FreeMorph & \textbf{0.335} & 0.030 & 0.119 & 0.197 \\
 & KontinuousKontext & 0.748 & 0.046 & 0.072 & 0.131 \\
 & VeloEdit & 0.577 & \textbf{0.061} & \textbf{0.053} & \textbf{0.097} \\
\midrule
\multirow{3}{*}{Change Attr. Color}
 & FreeMorph & 0.378 & 0.295 & 0.153 & 0.244 \\
 & KontinuousKontext & 0.084 & 0.548 & 0.104 & 0.177 \\
 & VeloEdit & \textbf{0.063} & \textbf{0.565} & \textbf{0.086} & \textbf{0.136} \\
\midrule
\multirow{3}{*}{Change Attr. Material}
 & FreeMorph & 0.311 & 0.087 & 0.151 & 0.234 \\
 & KontinuousKontext & \textbf{0.140} & 0.159 & 0.090 & 0.153 \\
 & VeloEdit & 0.183 & \textbf{0.179} & \textbf{0.079} & \textbf{0.131} \\
\midrule
\multirow{3}{*}{Change Background}
 & FreeMorph & 0.399 & 0.151 & 0.254 & 0.346 \\
 & KontinuousKontext & 0.203 & 0.208 & 0.159 & 0.224 \\
 & VeloEdit &\textbf{ 0.170} & \textbf{0.265} & \textbf{0.150} & \textbf{0.202} \\
\midrule
\multirow{3}{*}{Change Style}
 & FreeMorph & 0.315 & 0.099 & 0.005 & 0.008 \\
 & KontinuousKontext & \textbf{0.017} & 0.129 & \textbf{0.003} & \textbf{0.005} \\
 & VeloEdit & 0.062 & \textbf{0.218} & \textbf{0.003} & \textbf{0.005} \\
\bottomrule
\end{tabular}
}
\end{table*}

\begin{table}[tbp]
\centering
\caption{Analysis of computational overhead and efficiency. We report the total inference time for 100 images alongside the additional intervention cost introduced by VeloEdit. The marginal overhead confirms the efficiency of VeloEdit.}
\label{tab:time_compare}
\begin{tabular}{lccc}
\hline
Method & Total time (s) & Intervention time (s) & Rate (\%)\\
\hline
$\text{VeloEdit}_{\text{Flux.1 Kontext}}$  & 210.9 & 0.031  & 0.015\\
\arrayrulecolor{gray!60} 
\cmidrule(lr){1-4}
\arrayrulecolor{black}
$\text{VeloEdit}_{\text{Qwen-Image-Edit}}$ & 737.3 & 0.041  & 0.005  \\
\hline
\end{tabular}
\end{table}

\subsection{Additional Ablation Studies}
\label{sec:app_ablation}

\begin{figure}[btp] 
    \centering 
    \includegraphics[width=0.98\textwidth]{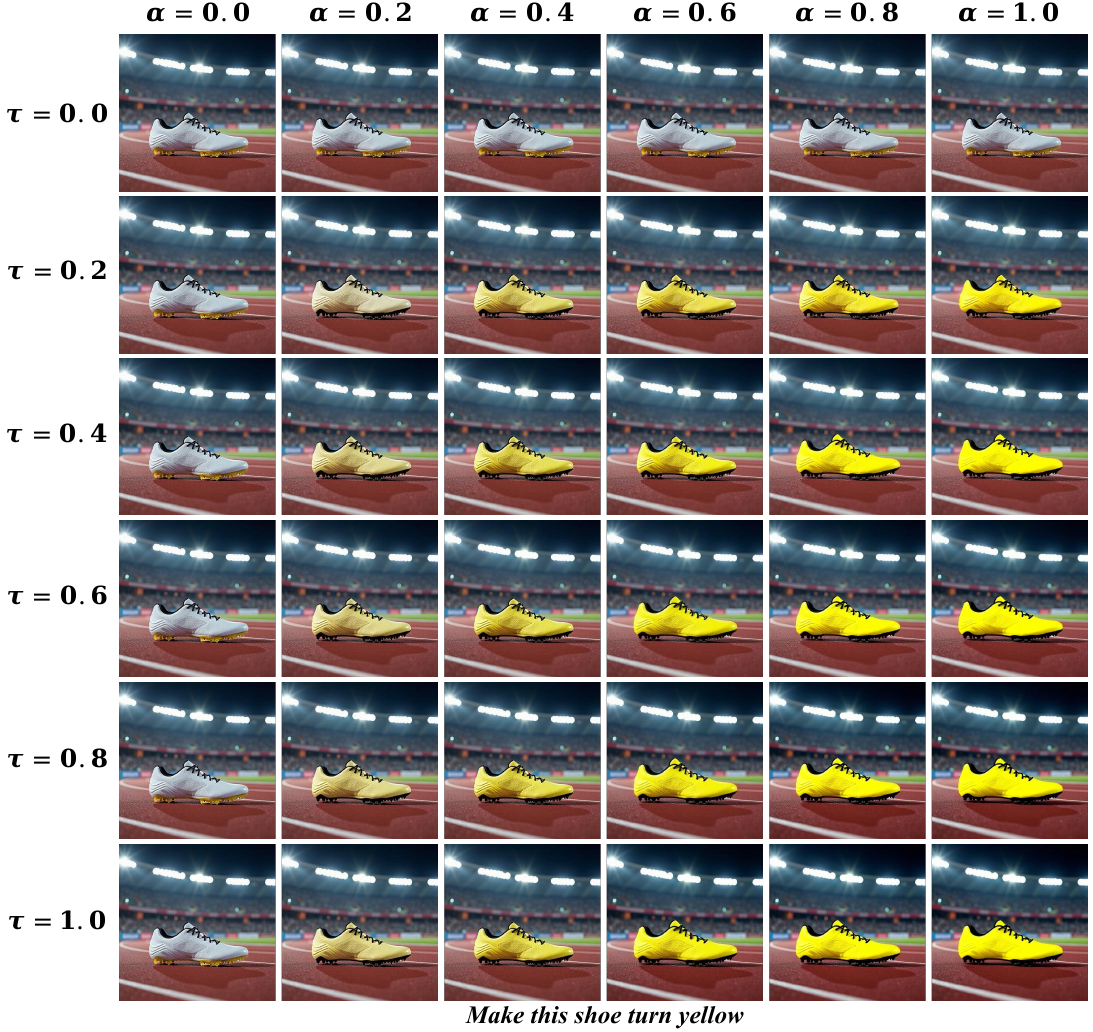}
    \caption{Visualizations of the ablation study on $\tau$. We illustrate the impact of varying $\tau$ on the continuity of the generated editing trajectories.} 
    \label{fig:abs_tau_1} 
\end{figure}

\begin{figure}[btp] 
    \centering 
    \includegraphics[width=0.98\textwidth]{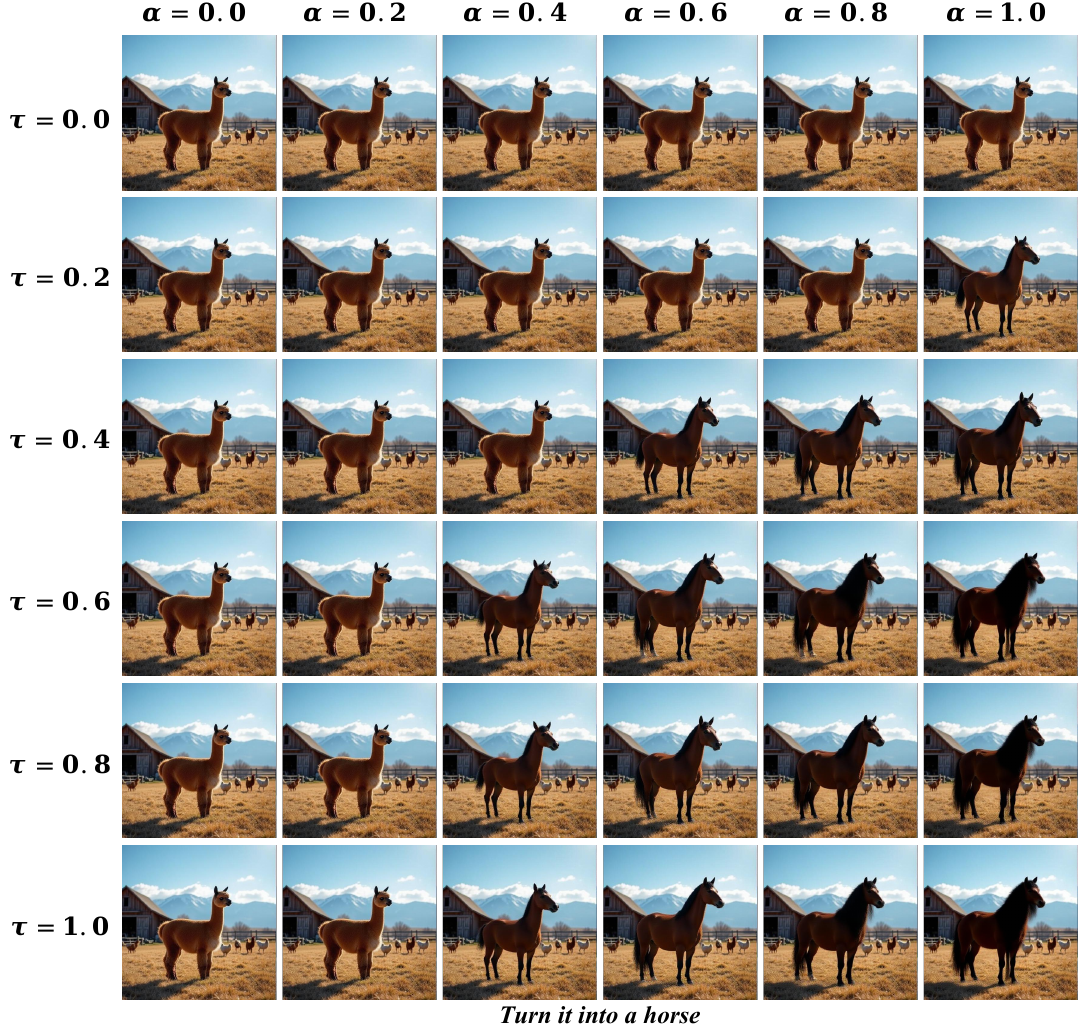}
    \caption{Visualizations of the ablation study on $\tau$. We illustrate the impact of varying $\tau$ on the continuity of the generated editing trajectories.} 
    \label{fig:abs_tau_2} 
\end{figure}

\begin{figure}[btp] 
    \centering 
    \includegraphics[width=0.98\textwidth]{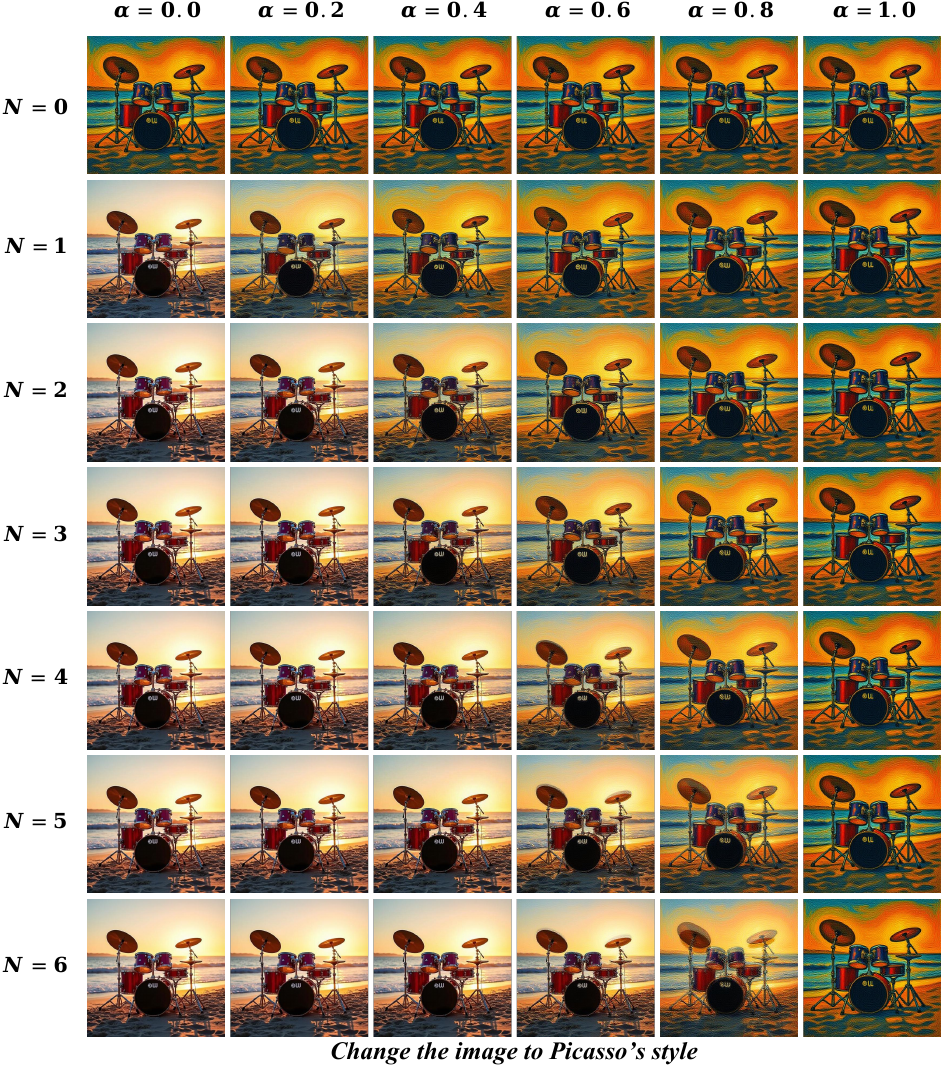}
    \caption{Visualizations of the ablation study on $N$. We illustrate the impact of varying $N$ on the continuity of the generated editing trajectories.} 
    \label{fig:abs_N_1} 
\end{figure}

\begin{figure}[btp] 
    \centering 
    \includegraphics[width=0.98\textwidth]{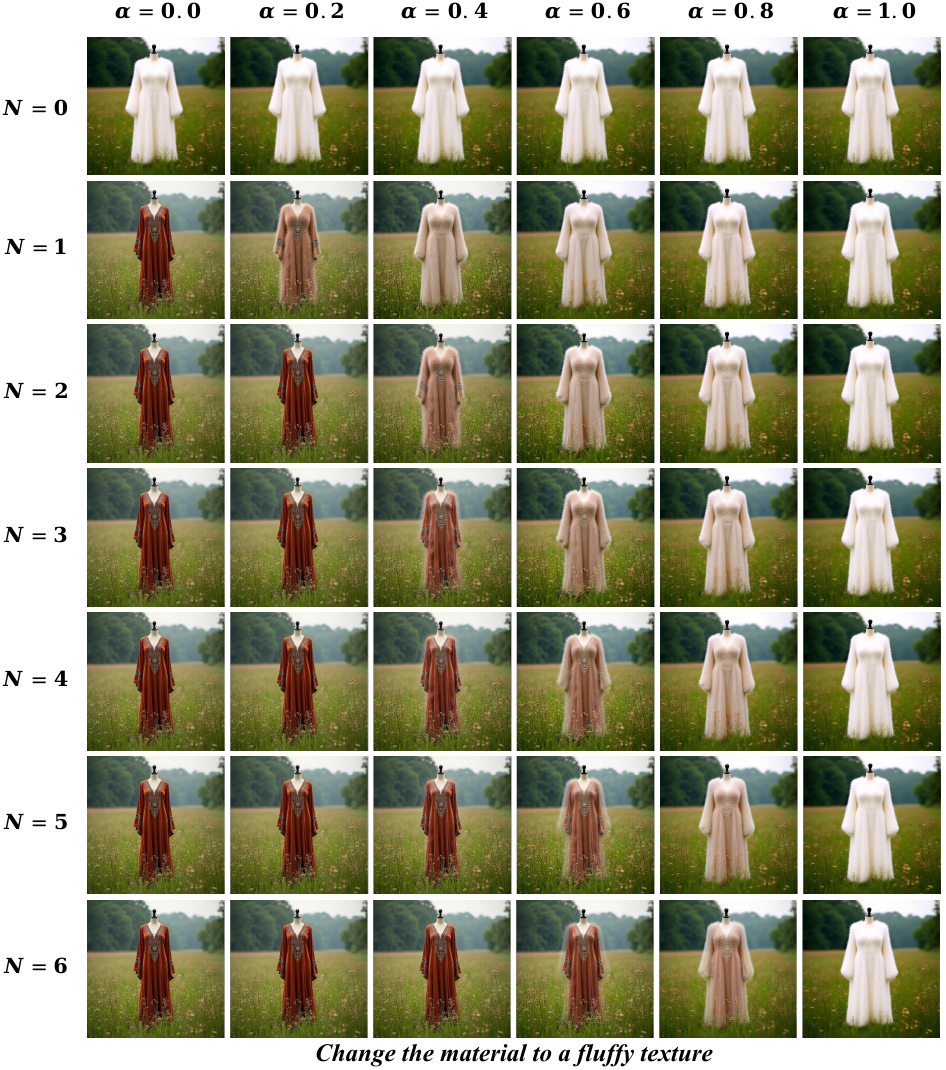}
    \caption{Visualizations of the ablation study on $N$. We illustrate the impact of varying $N$ on the continuity of the generated editing trajectories.} 
    \label{fig:abs_N_2} 
\end{figure}

In this section, we present the visual results of the ablation studies for $\tau$ and $N$, as shown in \cref{fig:abs_tau_1,fig:abs_tau_2,fig:abs_N_1,fig:abs_N_2}.

\subsection{Failure Cases}
\label{sec:app_imp_fail}

\begin{figure}[tp] 
    \centering
    \includegraphics[width=0.75\textwidth]{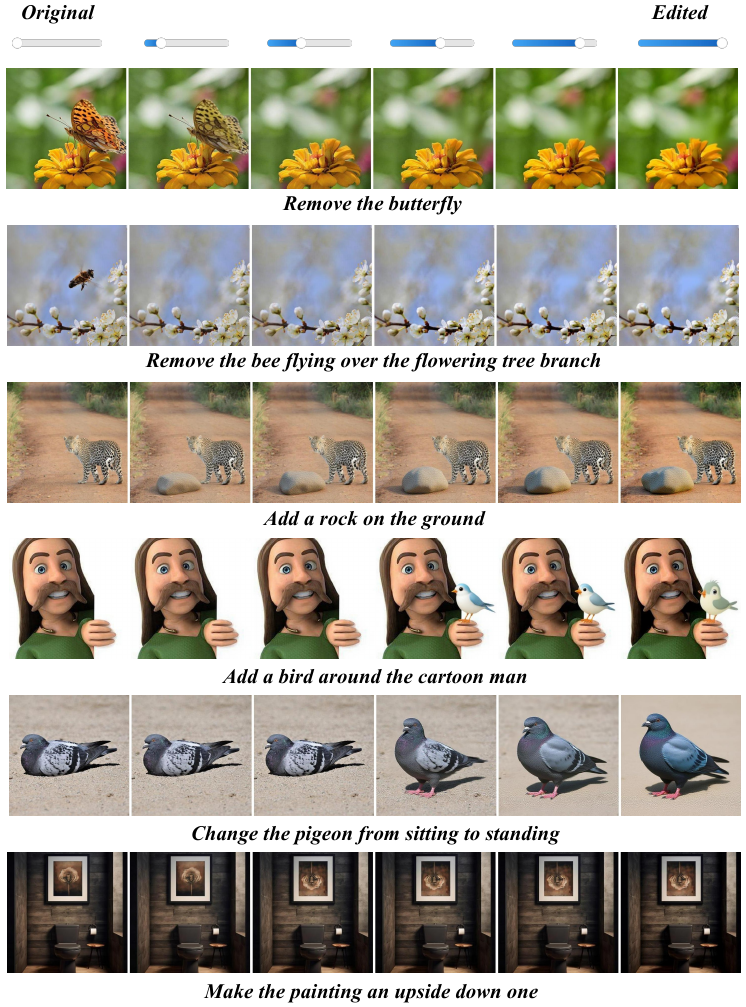}
    \caption{Failure cases. VeloEdit exhibits limited continuity in tasks involving object addition, object removal, and pose variation.} 
    \label{fig:add_f_7} 
\end{figure}

Similar to other continuous editing methods, VeloEdit struggles with tasks such as object addition, object removal, and significant pose variations. In these scenarios, it is prone to artifacts, abrupt changes, or meaningless outputs, as detailed in \cref{fig:add_f_7,tab:benchmark_comparison}.

\end{document}